# Towards Fleet-wide Sharing of Wind Turbine Condition Information through Privacy-preserving Federated Learning

L. Jenkel[a], S. Jonas[a,b], A. Meyer[a]

[a] Bern University of Applied Sciences, School of Engineering and Computer Science, Biel, Switzerland
[b] Università della Svizzera italiana, Faculty of Informatics, Lugano-Viganello, Switzerland
Corresponding author: stefan.jonas@bfh.ch

**Abstract.** Terabytes of data are collected by wind turbine manufacturers from their fleets every day. And yet, a lack of data access and sharing impedes exploiting the full potential of the data. We present a distributed machine learning approach that preserves the data privacy by leaving the data on the wind turbines while still enabling fleet-wide learning on those local data. We show that through federated fleet-wide learning, turbines with little or no representative training data can benefit from more accurate normal behavior models. Customizing the global federated model to individual turbines yields the highest fault detection accuracy in cases where the monitored target variable is distributed heterogeneously across the fleet. We demonstrate this for bearing temperatures, a target variable whose normal behavior can vary widely depending on the turbine. We show that no turbine experiences a loss in model performance from participating in the federated learning process, resulting in superior performance of the federated learning strategy in our case studies. The distributed learning increases the normal behavior model training times by about a factor of ten due to increased communication overhead and slower model convergence.

## 1. Introduction

Wind energy plays a pivotal role in climate change mitigation. A massive growth in the installed wind power capacity and grid infrastructure is required to decarbonize the power supply (Barthelmie & Pryor, 2021; Edenhofer et al., 2011). New wind farms are being planned and commissioned on an unprecedented scale in many countries (IEA, 2021, 2022; OECD et al., 2018). Access to the condition monitoring information from wind farms is an important requirement for reducing downtimes and enabling condition-based maintenance of wind farms (Carroll et al., 2016; Faulstich et al., 2011; Pinar Pérez et al., 2013). Despite this, manufacturers have been guarding condition data and reliability information from their turbines and are reluctant to share them due to business strategic interests (Kusiak, 2016). A strong data lack has resulted (Clifton et al., 2022; Leahy et al., 2019) which has been hampering the development, large-scale validation and operational deployment of data-driven models for wind turbine monitoring and diagnostic tasks.

Our study addresses this problem by proposing a privacy-preserving approach for sharing wind turbine (WT) condition information within a fleet of WTs of different owners without sharing any data from the WTs. In the context of this study, a fleet is the set of all WTs of the same model. The WTs within a fleet are identical in design. We demonstrate how data-driven condition monitoring models can be trained collaboratively by a WT fleet in a manner that allows sharing condition information among the WTs without sharing the WTs' condition data. Specifically, we propose to train accurate turbine-specific models of each WT's normal operation behavior for fault detection tasks by making use of the condition monitoring data of the entire WT fleet in a privacy-preserving manner. This is a highly relevant scenario because in practice only the manufacturer can access the condition data from all WTs of a fleet, whereas other stakeholders only have access to the small share of the fleet's data from their own WTs or even to no data at all (Kusiak, 2016). Other stakeholder groups concerned may include operators, owners, third-party companies, regulators, and researchers.

Thus, our study demonstrates a path towards privacy-preserving sharing of condition information without any manufacturer, operator or owner having to grant anyone access to their WTs' operation and condition data. Wind farm operators usually have no access to WT data from other operators and are, therefore, not able to make use of condition information from other operators' WTs for their own wind farms. The lack of data sharing (among wind farms of different owners) within fleets is particularly unfortunate in situations where the relevant data are scarce: For example, when the operator or other stakeholders seek to establish a damage database but have only few (or even no) fault events of each fault type in their database, or when a new WT has been commissioned and the stakeholder has no condition data available yet for that WT type. In such situations, it would be highly desirable to benefit from fleet-wide information sharing. Manufacturers, on the other hand, usually have access to the operation and condition data of all operating WTs produced by them but do not share these data.

To address the data imbalance, we propose and investigate the potential of privacy-preserving federated learning (McMahan et al., 2017) for condition monitoring and diagnostics tasks in wind farms based on WT data distributed among multiple owners. Federated learning has received growing interest in the field of mobile devices and Internet-of-Things applications after McMahan et al. (2017) presented the FedAvg algorithm. There have been numerous recent improvements of and contributions towards FedAvg, for instance, in enhancing security and privacy (Mothukuri et al., 2021; Yin et al., 2022) and in improving its efficiency (Acar et al., 2021; Asad et al., 2020). Comprehensive reviews of federated learning algorithms have been provided in (Aledhari et al., 2020; Kairouz et al., 2021; L. Li et al., 2020; T. Li et al., 2020; Lim et al., 2020; Mothukuri et al., 2021; Yang et al., 2019). An application of federated learning that is in use operationally are next-word predictions for virtual keyboards in mobile apps (Hard et al., 2018; Pichai, 2019). First applications have also been proposed in other fields, such as automotive systems (Y. Liu et al., 2020; Lu et al., 2020; Thorgeirsson et al., 2021). The capabilities of federated learning are still largely unexplored in renewable energy applications. Recently, Zhang et al. (2021) proposed a federated learning case study for probabilistic solar irradiance forecasting. Their presented FedAvg-based framework, enhanced by secure aggregation with differential privacy, was shown to achieve performance advantages over a setting in which data sharing between participants was unavailable. However, the authors noted that the shared federated learning model resulted in slightly inferior performance compared to a centralized setting with data sharing, as it is susceptible to data distribution deviations between clients. Lin et al. (2022) presented a federated learning approach for community-level disaggregation of behind-the-meter photovoltaic power production. To address the data heterogeneity of each community, a layerwise aggregation was introduced. Only the parameters of the shallow layers, learning community-invariant features, were exchanged, while the community-specific parameters of the deep layers remained local. This customization step was shown to result in improvements compared to a completely shared global model. With a focus on efficiency, Q. Liu et al. (2022) demonstrated a successful federated learning application to collaborative fault diagnosis of photovoltaic stations. To address the inefficiencies of FedAvg, especially when computing capabilities and dataset sizes differ between the participants, the authors proposed asynchronous decentralized federated learning. This framework without a central server resulted in significant reductions in communication overhead and training time.

In the field of wind energy, Cheng et al. (2022) presented the first and, to our knowledge, so far only study of a federated learning model for wind farms. The authors proposed an approach for detecting blade icing by classification. A blockchain-based architecture with a cluster-based learning module was introduced to address concerns regarding privacy and malicious attacks, as well as the data imbalance. The authors remarked that, while not considered in their study, existing data heterogeneity may negatively affect the performance of the model.

Classification methods such as Cheng et al. (2022) are relatively uncommon though for fault detection tasks in wind farms in practice due to the typically small number (or even absence) of fault observations. In contrast, fault detection based on normal behavior models is more common because it relies on learning an accurate representation of only the normal behavior of the WT and does not require a comprehensive amount of representative fault condition data, unlike fault classification approaches (Tautz-Weinert & Watson, 2017). Normal behavior modelling involves modelling the behavior of the monitored WT under normal fault-free operation conditions. The resulting normal behavior models (NBMs) characterize the normal operation behavior of the monitored WT as expected under the prevailing operating conditions. For example, NBMs can predict bearing temperatures or the active power output expected under the current normal conditions. NBMs are very useful because they enable the detection of significant deviations from the normal behavior that may indicate operation faults and trigger further investigation (Bilendo, Badihi, et al., 2022; Bilendo, Meyer, et al., 2022; Schlechtingen et al., 2013a). Such deviations can be detected based on the residuals of the measured and the expected state. SCADA-based NBMs have been proposed for single and for multiple monitored state variables (Meyer, 2021; Schlechtingen et al., 2013b; Zaher et al., 2009).

Multiple sensor systems are usually available in a WT for condition monitoring and normal behavior modelling for fault detection and diagnosis. They include temperature sensors, accelerometers for monitoring the vibration responses in the drivetrain and tower, oil quality sensors, and environmental sensors such as anemometers (Badihi et al., 2022; García Márquez et al., 2012; Tchakoua et al., 2014; Wymore et al., 2015). Condition monitoring can also be performed based on data from the supervisory control and data acquisition (SCADA) system of the WT (e.g., (Dao, 2022; Tautz-Weinert & Watson, 2017; A. Wang et al., 2022; Zaher et al., 2009; Y. Zhu et al., 2022) and based on combinations of SCADA and vibration data (e.g., (Sun et al., 2022)). SCADA-based condition monitoring can be considered a low-cost approach since no additional sensor systems need to be installed. On the other hand, the WT health information provided by SCADA data may be less informative in that it can be less component-specific, less timely and less accurate with regard to the fault diagnostics task than dedicated sensing systems such as accelerometers. For example, gearbox faults can be identified from vibration measurements at an early stage of fault development (e.g., (Jonas et al., 2023)), whereas associated SCADA data, such as from the gearbox temperature, would allow the fault to be detected only once it resulted in an unusual increase of the gearbox temperature, i.e., at a late development stage. Such temperature increases typically result from abnormal heat generation that can originate from excessive friction. Therefore, in SCADA-based fault detection, a fault can often be detected only at a relatively advanced fault development stage in which initial damage may have already occurred. Nevertheless, SCADA-based fault detection is a popular monitoring technique due to its simplicity, low cost and complementarity to other condition monitoring techniques in WTs. Comprehensive reviews of data-driven approaches in condition monitoring and diagnostics for wind farms were provided by (Black et al., 2021; Nunes et al., 2021; Pandit et al., 2023; Stetco et al., 2019; Tautz-Weinert & Watson, 2017).

The potential of collaborative fleet-wide learning of normal behavior models for fault detection tasks in WTs based on SCADA data has not been discussed or investigated, despite its high relevance for practical applications. Our study addresses this research gap by proposing federated learning of normal behavior models in a data-privacy-preserving manner. We propose a solution to an important practical problem in wind farm monitoring and diagnostics: How to train NBMs for detecting developing faults in WT subsystems when SCADA and sensor data for training NBMs are missing or not representative of the WT's current operation. This is a major challenge in newly commissioned wind turbines and in turbines whose operation behavior changed, for example, due to large hardware or software updates. We demonstrate the federated learning of NBMs in two case studies for gear bearing temperatures and power curves in two wind farms.

The main contributions of our study are:
1. a new privacy-preserving approach to wind turbine condition monitoring,
2. a customization approach to tailor the federated model to individual WTs if the target variable distributions deviate across the WTs participating in the federated training,
3. federated training and customization are demonstrated in condition monitoring of bearing temperatures and active power.

Our study contributes to resolving a major problem: the "lack of data sharing in the renewable-energy industry [which] is hindering technical progress and squandering opportunities for improving the efficiency of energy markets" (Kusiak, 2016).

This study is structured as follows. Section 2 details our proposition for collaborative privacy-preserving learning for condition monitoring and diagnostics tasks in WT fleets. Section 3 presents two case studies of a federated learning of normal behavior models in bearing temperatures and active power. We report and discuss our results in section 4. Section 5 summarizes the conclusions from our study.

## 2. Federated learning of wind turbine conditions
### 2.1 Federated learning

In conventional machine learning, all data on which a model is trained need to be available and accessible in a central system. If the data belong to different owners, such a centralized setting requires that the data owners give up their data privacy by sharing their data with others. In contrast, federated learning is a machine learning approach that learns a task from the joint data of different data owners without disclosing the data or sacrificing their privacy. In a federated learning environment, multiple industrial systems (clients, in our case: wind turbines) train a machine learning model in a collaborative distributed manner such that each client's training data remain on its local client system, thereby preserving the privacy of the training data (McMahan et al., 2017; Smith et al., 2017). With federated learning, the training data are distributed across multiple client systems and are not located in one central system, as is the case with conventional machine learning. The parameters of a collaboratively trained model are learned from the distributed data without exchanging the training data among the client systems or transmitting them to a central system. Only updates of the locally computed model parameters are shared with and aggregated by the central system. The model training is collaborative in the sense that each client contributes to the joint model training task by using its locally stored data for that task.

We adopt the FedAvg federated learning approach of McMahan et al. (2017) in our study. For a formal definition, it is assumed that a fixed number of $J$ client WTs are participating in the federated learning process. Each client WT $j$ has a fixed dataset $D_j$ of size $n_j = |D_j|$. In our case study, this is the dataset from the SCADA system used for training a normal behavior model of the WT normal operation behavior. Each dataset $D_j$ is stored locally in the client WT and not accessible to other client WTs or the central system. The FedAvg training proceeds in iteration rounds, at the start of which a central server transmits the initial model parameters of the current round to the $J$ client WTs (Table 1). Then, each client WT $j$ updates the received model parameters by training on its local dataset $D_j$, and then transmits the update to the central server. The server updates the parameters of the global model by aggregating the updates received from all client WTs by averaging. The objective of the iterative FedAvg training process is to arrive at model parameters $w$ that minimize the sum of prediction losses $\mathcal{L}_i$ from the $J$ client WTs on all data points $(x_i, y_i)$ of their local datasets $D_j$,

$$\min_w \sum_{j=1}^{J} \sum_{i=1}^{n_j} \frac{n_j}{n} \mathcal{L}_i(x_i, y_i, w) \qquad (1)$$

In our case study, the model parameters $w$ will be weights of a feed-forward neural network. We compute the prediction losses $\mathcal{L}_i$ in terms of the mean squared errors. In each algorithm round $t$, the update step involves that the $J$ client WTs perform local weight updates in parallel, so each client WT performs a gradient descent step on its local data,

$$w_{t+1}^j = w_t^j - \eta \sum_{i=1}^{n_j} \frac{n_j}{n} \nabla \mathcal{L}_i(x_i, y_i, w_t^j) \quad \forall j = 1, \cdots, J \qquad (2)$$

wherein $\eta$ is the learning rate and $\nabla \mathcal{L}_i(x_i, y_i, w_t^j), i = 1, \cdots, n_j$ denotes the gradients on $D_j$ of client WT $j$ with regard to the model weights $w_t$. The central server then aggregates the received weights and returns an updated model state $w_{t+1} = \sum_{j=1}^{J} \frac{n_j}{n} w_{t+1}^j$ to the client WTs, which ends the current training round. The training steps are being repeated until a predefined stopping criterion is satisfied. The overall training process is summarized in Table 1.

**Table 1.** Steps in the training of a federated learning model based on McMahan et al., 2017.

> **Federated model training process**
> The server selects the model architecture and initial weights, and iterates the steps below.
>
> 1. The clients participating in the training are selected.
> 2. The central server transmits the starting weights of the current iteration to the participating clients.
> 3. Each client trains a local model with stochastic gradient descent on its local training data. The local training is performed by all clients in parallel.
> 4. The updated weights are transmitted to the server which averages them. The resulting average weights are sent to the clients to serve as their new model weights and starting weights of the next iteration.

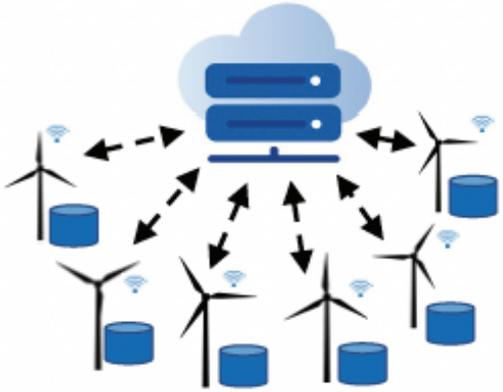

In addition to preserving the data privacy, further advantages of federated learning result from the fact that it does not require all client data to be stored in a central location. This can be highly beneficial when applied to complex remotely monitored power infrastructure such as wind farms. Modern WTs are equipped with hundreds of sensors that can collect hundreds of gigabytes of data every day (Siemens Gamesa, 2022). Transmitting and storing all those data in a central system (as would be common in conventional machine learning) is expensive and requires a high transmission bandwidth and data buffer. If the data are stored centrally, the data center managers of the central storage system are also responsible for protecting the data privacy and preventing unwanted third-party access, which entails an additional burden.

## 2.2 Federated learning for condition monitoring

Condition monitoring of wind turbines is often based on normal behavior models in practice (Schlechtingen et al., 2013b; Tautz-Weinert & Watson, 2017). Normal behavior models (NBMs) can be used for applications in fault detection and diagnostics. We propose and demonstrate the federated learning of normal behaviour models for such condition monitoring tasks. In the following, we analyze how NBMs for condition monitoring can be trained collaboratively by a fleet of WTs, in a manner that allows information sharing within the WT fleet without disclosing the data of any of the WTs. NBMs have been proposed for fault detection tasks based on SCADA and sensor data (Meyer, 2021; Schlechtingen et al., 2013b; Zaher et al., 2009). Our case studies explore the application of the FedAvg method (McMahan et al., 2017) for training accurate NBMs for fault detection applications in WTs which have few or no representative data. We focus on fault detection based on NBMs of drivetrain component temperatures and on the active power production (Meyer & Brodbeck, 2020). The drivetrain component temperatures exhibit more heterogeneous distributions across WTs. We investigate how federated learning can still be applied to extract accurate NBMs for condition monitoring, despite significant inter-turbine differences in the distribution of the target variable, in our case: the gear bearing temperature.

The temperature behavior of components and the active power form the basis of NBMs that are key for the condition monitoring in WTs (Kusiak et al., 2009; Lydia et al., 2014; Marvuglia & Messineo, 2012; Schlechtingen et al., 2013a; Shokrzadeh et al., 2014; Y. Wang et al., 2019). We demonstrate federated learning for NBMs of these applications.

Policies involved in the practical implementation of a federated learning process are beyond the scope of this study. There is certainly more than one setup and distribution of roles in the federated training process that can work in practice. For example, the federated learning process can be orchestrated by a regulatory entity who might define the process, the machine learning model structure, the aggregation, and distribute the software needed for the implementation. Federated learning can be organized in a centralized way, as presented here, but also in decentralized ways. Federated learning processes can be orchestrated by a central agency, such as a regulator. They may also be implemented and orchestrated by operators to enable data access across the fleet. Federated learning can, in principle, even be implemented by the manufacturer for customers who prefer not to give the manufacturer access to their turbines' data. In the centralized learning process proposed in our study, the client WTs only need to be equipped with a client computer that can train neural networks on their local data, with computing capacity and storage similar to that of a laptop computer.

**2.3 Customizing federated models to individual WTs**

A possible limitation of global federated learning models is that a single global model is trained for application in all client WTs of the fleet. Having a single non-customized model for all fleet members can limit the model's performance in the fault detection task, especially in cases in which the client WTs' SCADA and sensor datasets follow somewhat different statistical distributions in normal operation, requiring NBMs that are customized to each WT. Previous research investigating the effects of non-identically distributed data on the FedAvg algorithm has shown that data distribution differences can negatively impact the convergence and the performance of the global FedAvg model (Q. Li et al., 2022; Zhao et al., 2018; H. Zhu et al., 2021). We investigate NBM customization in our case studies.

The NBM resulting from the federated training process (Table 1) is a global model trained on the data of all client WTs, so it is not customized to a specific client WT. We demonstrate the limitations of a single non-customized model in our case studies based on the example of WT gear bearing temperatures and active power. Despite all WTs being the same model, each WT's local dataset can present a somewhat different data distribution. The arising data heterogeneity can be described as domain shift (Huang et al., 2023; Kouw & Loog, 2018; Quinonero-Candela et al., 2008), where the WTs' local datasets form diverse domains with different feature distributions. For example, the temperature behavior of the gear bearing can differ across WTs because of differing thermal behaviors. A single global NBM without customization learned through the FedAvg training process can lead to poor generalizability across domains (i.e., WTs) and to situations where for some client WTs the global NBM outperforms a locally trained one, whereas for other client WTs the global NBM performs worse than a model trained only on their local data. A lack of generalizability can become especially critical when WTs that have little or no representative data are dependent on information contained in the data of other WTs with distinct domains. Shared global models may be inadequate under these circumstances. Customized federated learning aims at alleviating this issue by customizing the global model to each client WT, while still participating in the distributed learning process. Customization techniques that have been proposed for federated learning models range from customization layers in neural networks (Arivazhagan et al., 2019) to meta-learning with hypernetworks (Shamsian et al., 2021). We refer to Kulkarni et al. (2020) and Tan et al. (2022) for an overview and taxonomy of customization techniques. In this study, we customize the global FedAvg model by means of local finetuning updates (Collins et al., 2022; Tan et al., 2022) which ensures that the participating client WTs can benefit from the federated learning process.

**3. Case studies: Federated learning of fault detection models**

The goal of our case studies is to estimate WT-specific normal behavior models for WTs that lack representative observations, and to perform the estimation in a collaborative privacy-preserving manner. A WT can suffer from a lack of representative training data for various reasons. A lack of representative data arises at the commissioning of a WT but can also occur after events that can affect the WT's normal operation behavior, such as control software updates or hardware replacements.

In the first case study, normal behavior models of the active power are developed: Some of the WTs participating in the federated learning process have representative local training data covering all wind conditions, whereas the training data of other WTs are dominated by low wind speeds. The second case study focuses on federated learning of normal behavior models of bearing temperatures. Unlike in the first case study, the bearing temperatures exhibit heterogeneous distributions across the WTs participating in the federated training. We show that customizing the trained global model to individual WTs yields the highest fault detection accuracy under such conditions.

The case studies are performed with data from two wind farms. The two wind farms are in separate locations (with a distance of at least 900 km) with different geographical and environmental factors. In the following sections 3 and 4, we will describe, present, and discuss our case studies with regards to data from the first wind farm. We then apply the same case study design and validate our results on the dataset from the second wind farm, which is presented in appendix A4.

SCADA data from ten commercial onshore wind turbines are analyzed for the case studies. All ten WTs are of the same manufacturer and model. The WTs are a horizontal-axis variable-speed model with pitch control and share the same technical specifications (Table 2). The data were acquired from the WTs' SCADA systems at a sampling rate of ten minutes over the course of 13 months. Each WT holds around 50'000 valid data points that contain wind speeds measured at the nacelle, the corresponding power generation, measured rotor speeds, and gear bearing temperatures. The measurements are provided as average values over 10-minute periods. All WTs are from the same wind farm, and we assume that no data sharing is allowed between the WTs. One randomly selected turbine out of the ten WTs is used only to define the network architecture with optimal hyperparameters, as explained in Appendix A1. The NBMs of the remaining nine client WTs are estimated based on the SCADA data.

**Table 2.** Technical specifications of the wind turbines employed in the case studies.

| Parameter | Specification |
|---|---|
| Rotor diameter | 112 m |
| Rated active power | 3300 kW |
| Cut-in wind velocity | 3 m/s |
| Cut-out wind velocity | 25 m/s |
| Tower | Steel monopole |
| Control type | Pitch-controlled variable velocity |
| Gearbox | Two planetary stages, one helical stage |

### 3.1 Federated learning of Active Power models

The first case study demonstrates the privacy-preserving collaborative learning of NBMs of a wind turbine's active power generation. The trained NBMs enable the detection of underperformance faults in the monitored WTs. The normalized 10-minute average wind speed serves as regressor in the normal behavior model of the power generation. The wind speed was min-max normalized such that all normalized wind speeds are in the range of [0,1]. We investigate a scenario in which five of the nine WTs are affected by a lack of representative SCADA data in the sense that the data are dominated by low and moderate wind speed observations whereas observations from time periods of high wind speeds are lacking. In practice, this scenario may arise when the existing WT data were taken during extended periods of low wind speeds which are not uncommon in many regions, e.g., in central Europe in summertime, and can last for several weeks and even months (Ohlendorf & Schill, 2020). This case study demonstrates just one of several possible scenarios of lacking representative training data for NBMs. It serves to illustrate the advantages of federated learning approaches for condition monitoring and fault diagnostics tasks.

A lack of representative SCADA data from a particular WT means that accurate NBMs can hardly be estimated for that WT with conventional machine learning approaches. It may take up to several months of SCADA data collection until a sufficiently representative dataset has been collected for training a new NBM from the WT's own SCADA data. Active power monitoring and detection of underperformance faults are hardly possible during this time period. We demonstrate that collaborative learning of the nine WTs can mitigate this lack of training data and allows learning accurate power curve NBMs in a privacy-preserving manner.

For each WT, we set aside the last 30% of its SCADA data, i.e., the data gathered in months ~9–13 of the 13-months data collection period, as that turbine's test set. Further, we assign nine randomly selected WTs as "client" turbines. The remaining WT is treated as a public turbine in the sense that its SCADA data will serve us for the model selection. The remaining 70% of the data of each client WT are split into a training set and a validation set in a manner that represents the data-scarce conditions as discussed above: We define the training set of each of the five WTs to be composed of the 10-minute average wind speed and power generation values of the four weeks with the lowest average wind speeds out of the considered 9-months measurement period. Thus, the training sets of the five WTs are characterized by low and moderate wind speed conditions. All other time periods form the validation set of that respective client WT. The training and validation set of the remaining four WTs comprise all wind resources, including low, moderate and high wind speeds. The last 30% of the training set data form the validation set for these clients. An illustration of the training, validation and test datasets is given in Figure 1 for one of the five data-scarce WTs and for one of the four WTs with representative training data. The accuracy of the power curves of the five WTs is limited due to the lack of observations of high wind speeds in their local training data.

Note that the data from the four WTs with representative training data are inaccessible to the data-scarce wind turbines. So it is not possible to derive and transfer a power curve from any of those four WTs to any data-scarce WT because the data are local and, thus, unavailable to standard (non-federated) learning approaches.

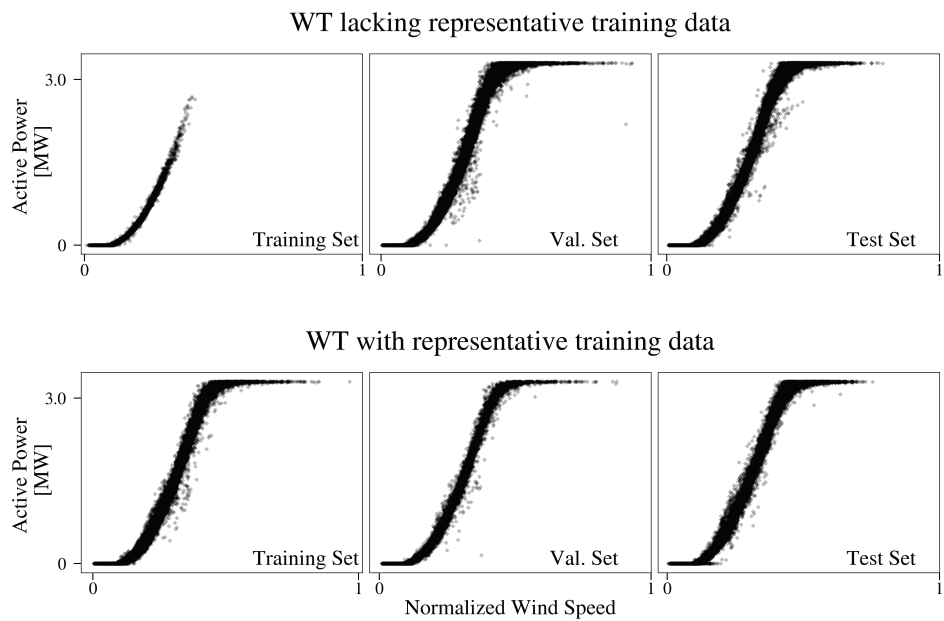

**Figure 1.** Datasets of two different client turbines. First row: Only data from the four weeks with the lowest average wind speed were kept for the training set of this client turbine. The training set does not contain sufficient data to represent the true power curve behavior in high wind speed situations (upper left panel). Second row: Wind speed and power data from a client WT whose training data contain representatively distributed wind speed observations.

### 3.2 Federated learning of Bearing Temperature models

Our second case study demonstrates the federated learning of NBMs of bearing temperatures for fault detection applications. Unusually high component temperatures can be caused, for example, by excessive friction or undesired electrical discharges, so excessive temperatures are key SCADA indicator variables of developing operation faults. The normal operation behavior of gear bearing temperatures is modeled with normalized 10-minute rotor speeds and power generation as regressor inputs. Again, five of the nine WTs are affected by a lack of representative SCADA data. Specifically, only one month of local training data is used to train the NBMs of these WTs. Such scarcity conditions regularly arise in newly installed WTs and after major software or hardware updates. As in the first case study, the last four months (30% of the SCADA data) serve as the WT's test set. Nine randomly selected WTs are assigned as client WTs, whereas the remaining WT is used for the model selection. The remaining 70% of the SCADA data of each client WT are split into a training set and a validation set in accordance with the data scarcity conditions: The training set of each of the five WTs is defined to be composed of the 10-minute average gear bearing temperatures, rotor speeds, and active power generation values of one month, i.e., four randomly chosen consecutive weeks. All other time periods form the validation set of the respective WT. The datasets of the remaining four WTs comprise a longer, more representative data collection, where the last 30% of the training set data form the validation set for these clients. Figure 2 illustrates the training, validation and test datasets for one of the five data-scarce WTs and for one of the four WTs with representative training data.

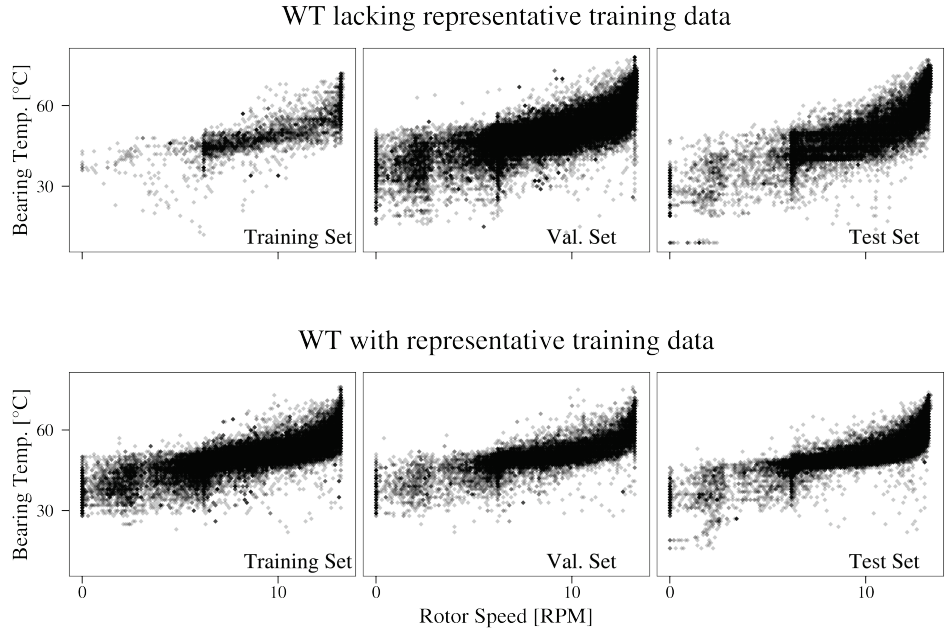

**Figure 2.** Datasets of two client WTs. First row: Only data from four randomly chosen consecutive weeks were kept for the training set of this client turbine. In this case, the training set contains insufficient data to represent the temperature behavior in low temperature situations (upper left panel). Second row: Gear bearing temperature and rotor speed data from a client WT whose training data contain representatively temperature observations.

### 3.3 Heterogeneously distributed target variables

Deviations in the data distributions across the local datasets of the participating client WTs can negatively affect the FedAvg learning process, as discussed in section 2. Our case studies exhibit different degrees of distributions shifts in the target variables, enabling us to investigate the effects of deviating distributions of the monitored variables. Figure 3 shows the distributions of the active power generation and the gear bearing temperatures across the nine WTs participating in the federated training. The distributions of active power (i.e., the target variable of the first case study) display only minimal differences across the client WTs. So one expects that a global federated model should be able to capture information that can be generalizable across WTs. We assess how this global knowledge can be shared and utilized by WTs with scarce training datasets in the case studies. We also evaluate how the loss of WT-specific information in the global model affects WTs with representative training datasets, and the utility of customized models under these conditions.

The distributions of the gear bearing temperatures show distinct differences across all client WTs. A globally shared model may have difficulties capturing global information that is generalizable across WTs. Customization to individual WTs may improve the model performance in the case of non-identically distributed datasets. We assess the effect of the observed distribution shifts on the performance of the global model in the second case study, and whether collaborative condition information sharing across the fleet is still possible and beneficial for the participating WTs under these conditions.

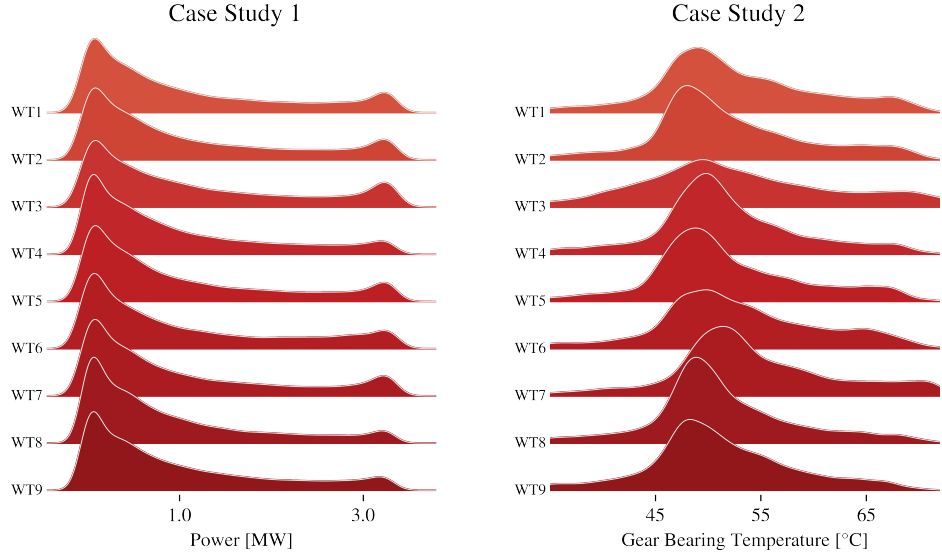

**Figure 3.** Kernel density estimates of the distributions of the monitored variables on the test set of all nine client WTs.

## 4. Results
### 4.1 Federated learning strategies and model architecture
The scarcity of training data is addressed by privacy-preserving information sharing between all WTs active in the federated training: The local data of each client WT contribute to training the global NBM and to fine-tuning it to the respective client WT. Yet, the local data remain stored in the respective client WT without exposing them to other client WTs or the central server in the federated training process (Table 1). We compare three privacy-preserving learning strategies (Figure 4):
  A. Conventional machine learning of NBMs using only the local data of each WT,
  B. Federated learning of a single global NBM for all WTs,
  C. Customized federated learning of WT-specific NBMs.

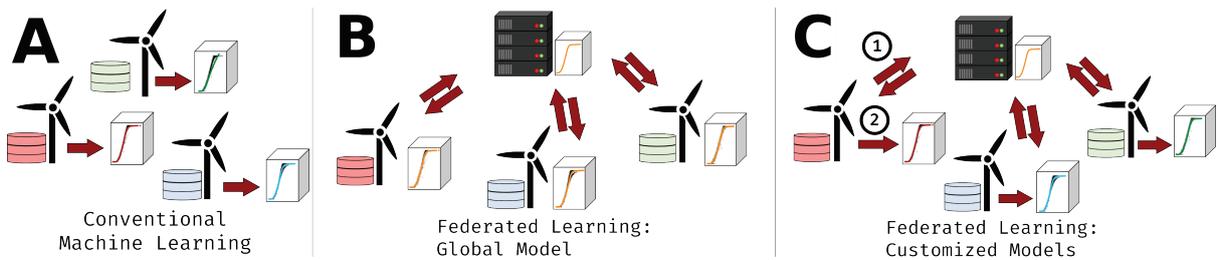

**Figure 4.** Learning strategies applied in the case studies: Conventional machine learning (A), Federated learning of a single global NBM (B), Customized federated learning of WT-specific NBMs (C).

**A. Conventional machine learning**
We evaluate the training of a NBM in a conventional non-distributed machine learning environment. Each client WT individually learns a NBM based on its own past operation data and without any access to data from other WTs of the fleet. This constitutes the default situation in practice. We typically lack access to data from other fleet members because they have other owners and no data sharing is in place.

**B. Federated learning of a single global model**
Our second training strategy for the NBM is a federated learning environment. In this setting, a central server communicates with the client WTs in a privacy-preserving manner. We implement the federated averaging approach of McMahan et al. (2017), see Table 1. First, the server broadcasts the model architecture, determined with the model search over the server-accessible public WT, and further information such as the optimizer, loss, and metrics to the client WTs in the initialization step. The iterative update step consists of the client WTs first updating their models in parallel – which we implemented as running three epochs over their private local training sets – and then sending their model weights back to the server. Next, the server averages the collected client weights and broadcasts the averaged model weights to the client WTs. The averaged model weights represent the global FedAvg model. An additional sidestep involves that all clients evaluate the updated global model on their validation set and send their validation losses to the server. We repeat the update step until the average validation loss of the clients has not improved within 5 repetitions, representing 15 local epochs by each client. The global federated learning model is then evaluated by calculating the root mean squared error (RMSE) on the test set of each client WT.

### C. Customized federated learning of turbine-specific models

A possible disadvantage of the presented federated learning approach (B) is that it results in a single global model that is not customized to a specific client WT. The individual client WTs may exhibit somewhat different data distribution characteristics depending, for example, on their sites, maintenance history, or local ambient conditions. The feature distributions of the monitored target variable may differ significantly across the fleet (Figure 3). Such differences are not represented by the global NBM, which may result in performance losses of the model for some client WTs. Some turbine operators might be incentivized to opt out of the federated learning process if they find that a local NBM trained only on their local data with conventional machine learning (A) outperforms the global NBM (B). Training WT-specific NBMs can make it attractive for all client WTs to join the training, so we customize the MLP that represents the global NBM to specific client WTs. After training a single global MLP model based on FedAvg (B), we achieve the customization by having each client WT finetune a subset of the trained layers of the global MLP on its local dataset (Figure 5 and appendix A3). This turbine-specific finetuning resembles transfer learning methods in which neural network layers of a previously trained model are finetuned on a separate dataset (Collins et al., 2022; Kulkarni et al., 2020; Pan & Yang, 2010; Tan et al., 2022; Zhuang et al., 2021). Based on the validation set losses, each client WT optimizes the number of layers to finetune in its customized MLP. The weights of the other layers remain fixed with the weights of the global federated learning model. The resulting model performances are presented in Appendix A3. The customized model with the lowest RMSE on the client WT's validation set was finally evaluated on each test set.

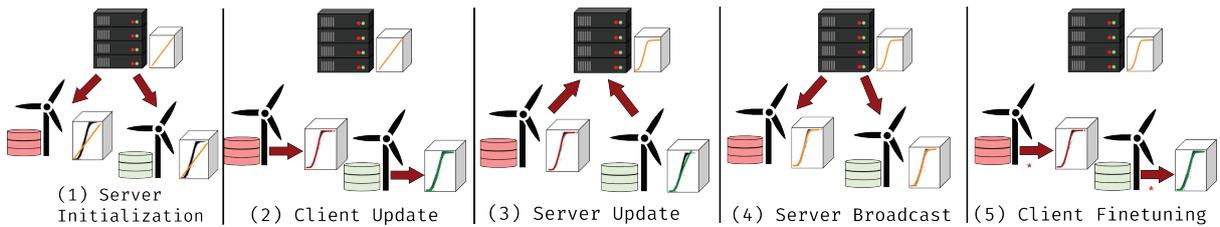

**Figure 5.** Illustration of the federated learning process with customization in step 5. Step 1: The server initializes an empty model and broadcasts the architecture to the clients. Step 2: Each client updates their model weights by running training epochs over their private local datasets. Step 3: The clients broadcast their model weights to the server which aggregates them into a server model. Step 4: The server broadcasts the calculated model to the clients. Steps (2)–(4) are repeated until a training stop criterion is satisfied. At the end of step 4, the server and clients share the same ("global") model weights. The customization step 5 involves the finetuning of layer weights of the global model trained in steps 1–4.

### 4.2 Federated learning of Active Power models

In case study 1, a feedforward neural network is trained as a NBM of the power generation. The inputs to the model are the normalized SCADA wind speeds. The model outputs a prediction of the active power generation in MW. In each case study, all client WTs and federated learning strategies make use of the same feedforward multilayer perceptron (MLP) model architecture to ensure fair comparisons among experiments. The respective MLP architecture will be determined by applying a random search model selection algorithm on the SCADA dataset of the public turbine. For the first case study, the resulting model architecture is summarized in Table 3. The search algorithm and model architecture selection are outlined in Appendix A1. Each client WT minimizes the mean squared error loss over the training set by applying stochastic gradient descent (SGD) in the conventional machine learning according to strategy A. Training is stopped once the client WT's validation set loss has not improved within 15 epochs. The model performance is finally evaluated as the RMSE on the test set of over the client WT. The results are summarized in Figure 6. Detailed results for each WT are presented in the Appendix A2.

**Table 3.** The model architecture of the Active Power NBM used in all experiments of the first case study.

| Model architecture (First case study) |
| --- |
| • Input layer: Normalized wind speed value |
| • Layer 1: Fully connected, 12 units, exponential linear unit activation function |
| • Layer 2: Fully connected, 8 units, exponential linear unit activation function |
| • Output layer: Fully connected, 1 unit, Rectified Linear Unit activation function (Nair & Hinton, 2010) |
| • Output: Predicted Power Value [MW] |
| |
| Number of parameters: 137 |
| Loss: Mean Squared Error |
| Optimizer: Stochastic Gradient Descent (learning rate = 0.013, batch size = 32) |

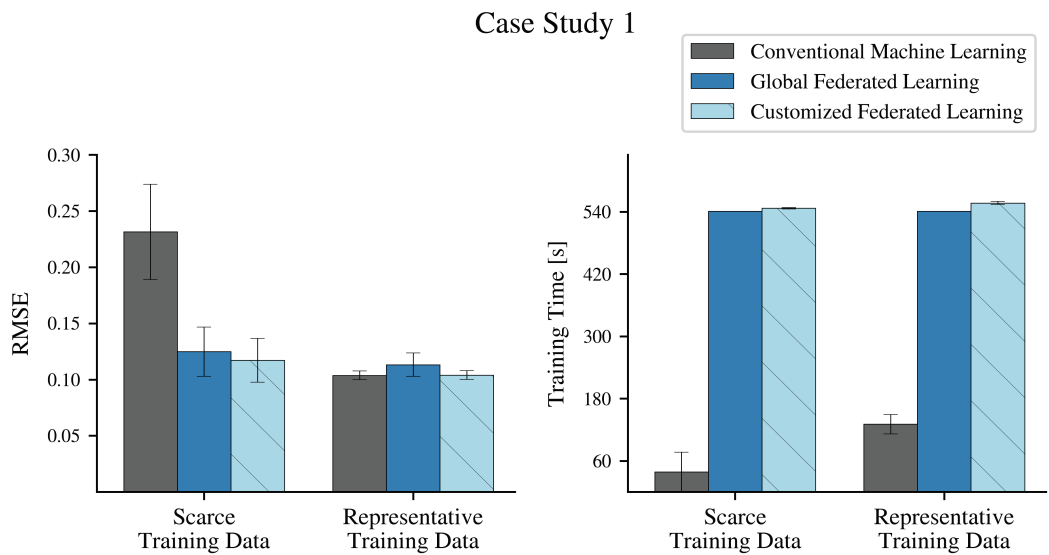

**Figure 6.** Left: Performances of the training strategies on the test set in terms of mean RMSE between the NBM predicted power and actual monitored variable in case study 1. Right: Mean training time in seconds for all three learning strategies. The error bars display the standard deviation.

**Model performance.** The performances of the three privacy-preserving NBM training strategies (A-C) are compared with regard to the accuracy of the resulting active power NBMs on the test sets of each of the nine client WTs and with regard to the model training time. In conventional machine learning (strategy A), we find a significant difference in model performance depending on the type of the training dataset. The client WTs trained on the four weeks with the lowest wind speed average of the considered 13-months period show a significantly higher error (mean: 0.231) than those WTs with training datasets of all wind speed conditions (mean: 0.104), as shown in Figure 6. Due to the scarcity of high-wind speed observations, conventional machine learning on the local client's data cannot train a sufficiently accurate power curve NBM. An example of such a client WT dataset is shown in Figure 7.1. The corresponding model trained with gradient descent on only the local training data (strategy A) does not capture the power curve behavior correctly at higher wind speeds (Figure 7.3). For the four WT clients with representative wind speed data, the power curve can be fit accurately even with conventional machine learning with only the local training data.

The results by the global federated learning model (strategy B) show a contrast between the client WTs with scarce high wind speed observations and the client WTs with representative wind speeds. For the WTs with scarce high wind speed observations, the RMSE of the active power NBM are significantly reduced by the global federated learning (mean: 0.125) compared to the conventional machine learning setting. By receiving shared model parameters from all client WTs through the server aggregation step, the client WTs with scarce high wind speed observations are now able to also model the upper wind speed ranges by means of the shared global model. Therefore, these client WTs benefit from the federated learning process through a significant improvement in model performance. Panel 7.4 shows the accordingly improved power curve of one of the five client WTs with few or no high wind speed data with a realistic behavior in the upper ranges, despite not having any reference data points available in its own local training set.

Conversely, the model performance has slightly but noticeably decreased for all but one of the four client WTs with representative wind speed observations (mean: 0.113) by the global federated learning as compared to the conventional machine learning setting. The averaging step of the global federated learning leads to a loss of individual characteristics contained in the local models of those client WTs. Therefore, as these clients were already locally capable of fitting a model tailored to their individual turbine characteristics, also in the upper wind speed ranges, the averaged global federated learning model leads to a performance loss by incorporating individual information from other turbines.

Such performance losses could discourage operators of client WTs with sufficiently representative training data from joining the federated learning process. These client WTs should not drop out of the federated learning though because they are essential to the performance increase of the client WTs with scarce data in this example. Indeed, our results show that a customized federated learning implementation can counteract this issue. The local finetuning of the global federated learning model manages to revert the impact of the global averaging and re-introduces individual characteristics into the models. Thus, the active power NBMs include both global information as well as customized adjustments. Panel 7.5 shows that the active power NBM from the customized federated learning model is very similar to but somewhat deviating from the global federated learning model to correct for local dataset characteristics.

Comparing the average performances of the three learning strategies (A–C), the customized federated learning approach (C) accomplished the lowest RMSEs for the clients with scarce high wind speed observations (mean: 0.117) and achieves the same performance as the conventional machine learning strategy (A, mean: 0.104) for clients with representative wind speed observations. Our results suggest that a customization method should be applied for possible performance improvements of the trained NBMs and as an incentive for all client WTs to join the federated training process.

Compared to conventional machine learning (A), a distributed learning process such as federated learning requires additional computational costs due to the communication between server and clients, overhead operations, and slower model convergence. Figure 6 shows the measured computational time taken to accomplish the training process for the three learning strategies. All client WTs finish training within less than three minutes in a conventional machine learning setting (A). With a global federated learning strategy (B), the clients require more than 9 minutes for the learning to be accomplished. Given this increase, the training time needs to be investigated when considering federated learning applications for more complex models and for training with a larger number of client WTs. In customized federated learning (C), the computational costs are dominated by the global learning step as the customization step requires a finished federated learning process. The added time taken by the actual customization step, i.e., the local finetuning, becomes negligible (on average +10.4 seconds) in comparison. Thus, a local finetuning step is a very cost-efficient improvement.

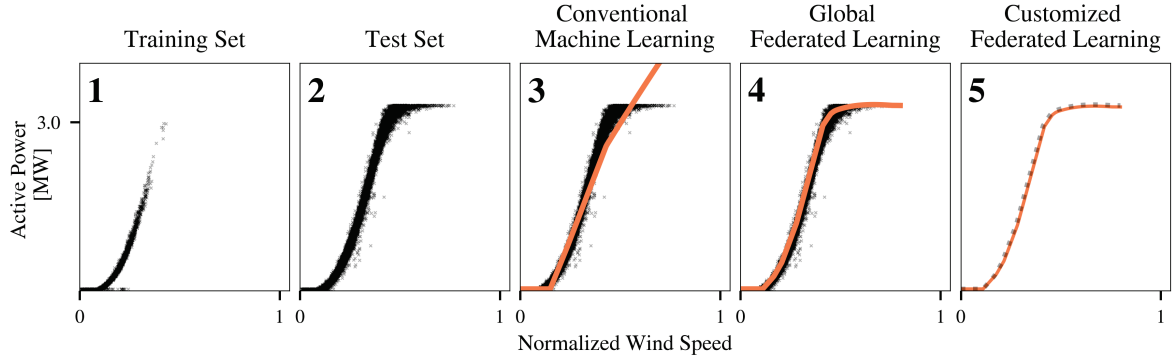

**Figure 7.** Training set (7.1) and test set (7.2) for a randomly selected one of the five WTs with few or no high wind speed data in their training sets, and the power curve models trained for that WT based on conventional machine learning (7.3), the global federated learning model (7.4) and the customized federated learning model (7.5). As the training set of the WT contains only few data points for high wind speeds, the conventional machine learning model fails at modeling the true power curve behavior for higher wind speeds, which is shown by the underlying test set data. By privacy-preserving learning from other WTs, the global federated learning model (7.4) can now model these higher ranges. The finetuning step in the customized approach slightly adjusts the global model (dashed line) to the private training set (7.5).

### 4.3 Federated learning of Bearing Temperature models

The privacy-preserving learning strategies A-C are also investigated in a second case study. A feedforward neural network is trained to model the normal behavior of the gear bearing temperature using SCADA data. To ensure fair comparisons between the strategies, all client WTs and federated learning strategies use the same feedforward MLP model architecture, outlined in Table 4.

**Table 4.** The model architecture of the Bearing Temperature NBM used in all experiments of case study 2, determined through a model search (appendix A1).

| **Model architecture (Second case study)** |
| --- |
| • Input layer: Normalized rotor speed and power |
| • Layer 1: Fully connected, 8 units, exponential linear unit activation function |
| • Layer 2: Fully connected, 16 units, exponential linear unit activation function |
| • Output layer: Fully connected, 1 unit, linear activation |
| • Output: Predicted Gear Bearing Temperature [°C] |
| Number of parameters: 185 |
| Loss: Mean Squared Error |
| Optimizer: Stochastic Gradient Descent (learning rate = 0.00035, batch size = 32) |

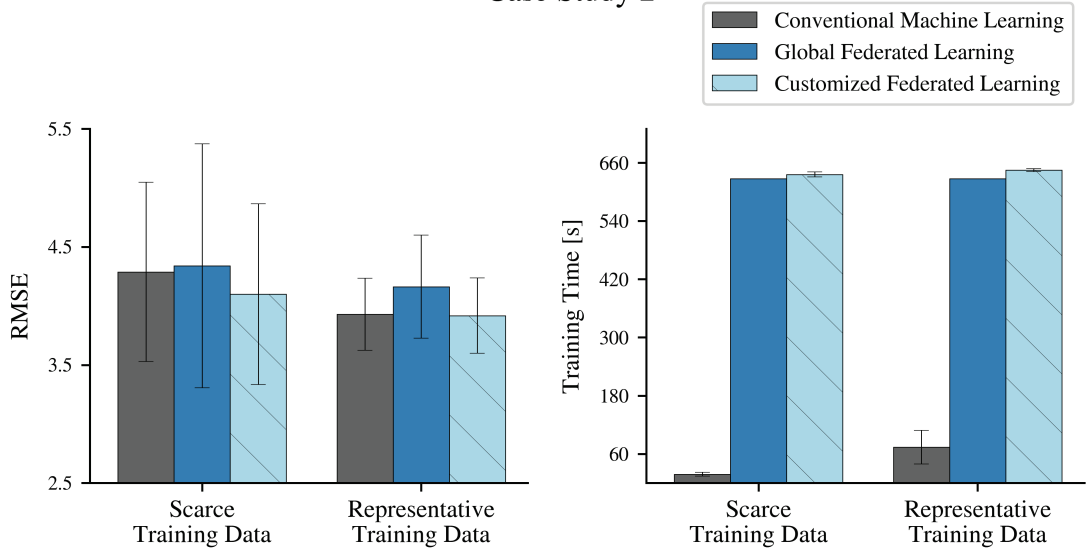

**Figure 8.** Left: Performances of the training strategies on the test set in terms of mean RMSE between the NBM predicted temperature and the actual monitored variable in case study 2. Right: Mean training time in seconds for all three learning strategies. The error bars display the standard deviation.

**Model performance**. The accuracies of the NBMs trained with non-collaborative strategy A are shown in Figure 8. WTs with scarce datasets (mean RMSE: 4.29) have a higher average RMSE than WTs with representative datasets (mean: 3.93) with this strategy. The models trained on scarce datasets with strategy A are not capable of fully capturing the temperature behavior. An example of this is shown in Figure 9, in which the trained model is unable to adequately estimate temperatures in underrepresented ranges (very low and high temperatures, as shown in 9.1), which leads to larger errors in lowest and highest observed temperature values on the unseen test dataset (9.2).

Comparing the performance of global federated learning (B) to conventional machine learning (A) for the five client WTs with scarce datasets, the global model leads to performance increases in only two client WTs but raises the prediction errors of the NBMs in three other WTs. The global models results in worse NBM performance even though the three WTs lack representative data and receive shared model parameters. This result suggests that the substantially differing bearing temperature behavior across clients strongly affects the generalizability of the global model, such that one global model trying to combine all individual characteristics cannot always offer a satisfactory fit. Therefore, despite receiving information about temperature ranges not represented in their training set, these values do not necessarily reflect the actual bearing temperature behavior of that WT. An example is shown in Figure 9.3 where the global model introduces a strong overestimation of the lower bearing temperatures.

For the four clients with a fully representative training set, the global federated learning model leads to a noticeable increase of the RMSE in all cases. The global model incorporates information from all turbines, leading to a loss of individual characteristics within the model and thus to a loss in performance, as already observed in case study 1.

A customized federated learning strategy can encourage operators of client WTs without data scarcity to participate in the federated learning process because a customized strategy can revert potential performance degradation introduced by the global model. Both case studies show that the customization step is a necessity to encourage clients without data scarcity to join the federated learning process. For clients with scarce datasets, the customized federated learning strategy achieves the best performance across all strategies. The local finetuning enables the customized models to retain and transfer usable knowledge from the global model (for data not represented in the scarce dataset) and additionally incorporate individual characteristics from the private local dataset. An example is shown in Figure 9.4 where the bearing temperature estimates from the customized model are now improved in the unseen low and high temperature ranges. Our results suggest that a customized federated learning strategy can enable fleet-wide learning of condition information even in the presence of a significant domain shift.

The computational times taken to train the NBMs following the three learning strategies (Figure 8) confirm the results of case study 1. We observe a strong increase in training time of the global federated learning model compared to conventional machine learning. Training a model according to the conventional machine learning strategy takes on average 43 seconds, while the federated learning process requires more than 10 minutes. In contrast, the increase in time for the local finetuning of the global model (customization part of strategy C) remains negligible as it only requires an additional 12.9 seconds of training on average. The results of case study 2 reinforce that a disadvantage of the federated learning process is its additional computational costs and that customized federated learning (strategy C) is a very time-efficient model improvement strategy. Detailed results for all WTs are shown in Appendix A2.

All experiments were run on an Intel Xeon CPU @ 2.20 GHz with implementations using TensorFlow v2.8.3, Keras v2.8, and the tensorflow-federated v.0.20.0 framework (Abadi et al., 2016; Chollet & others, 2015).

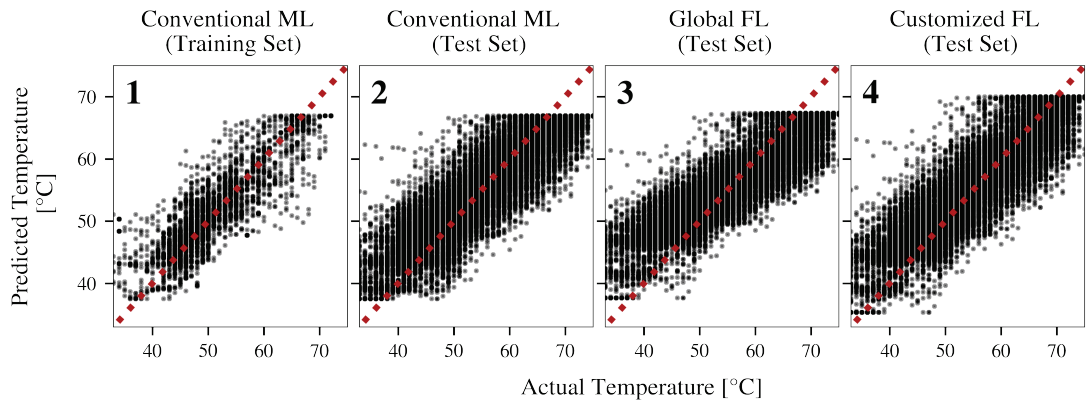

**Figure 9.** Actual versus predicted gear bearing temperatures based on a NBM of a WT with scarce training data. All data points would be located on the diagonal line with a perfect model. Panels 9.1 and 9.2 show predictions using conventional machine learning on the training set and test set, respectively. Panels 9.3 and 9.4 show the test set predictions by the global federated learning NBM and by the customized federated learning NBM.

### 4.4 Second wind farm

We further validate our findings by replicating our case studies using data from the second wind farm. The wind turbines in the two farms belong to different fleets. They have different manufacturers, different rated powers, and major constructional differences. Details and results are provided in appendix A4. The transfer across different fleets is not in the scope of our study.

### 5. Conclusions

A wealth of data is being constantly collected by manufacturers from their wind turbine fleets. Stakeholders interested in those data can include operators, owners, manufacturers, third-party companies, regulators, and researchers. There are various reasons why different stakeholder want access to information contained in a fleet's operation data. Benefits of making the information accessible include technological progress, for example through new and improved data-driven applications, and economic advantages resulting from increased transparency and competition. For example, improved machine learning models can be trained based on a fleet's data to provide better decision support to wind farm operators. This may involve improved predictions of failure events and estimations of the remaining useful lifetime of critical parts.

Conventional machine learning on local wind turbine datasets is often applied in practice but it cannot exploit the information contained in the operation data of distributed wind turbine fleets. Conventional machine learning cannot overcome the lacking access to fleet-wide data because it is incompatible with data privacy needs.

We have demonstrated a distributed machine learning approach that enables fleet-wide learning on locally stored data of other participants in the federated learning process, without sacrificing the privacy of those data. We have investigated the potential of federated learning in case studies in which a subset of wind turbines was affected by a lack of representative data in their training sets. The case studies involve the collaborative learning of normal behaviour models of bearing temperatures and power curves for condition monitoring and fault detection applications.

The results of our case studies suggest that a conventional machine learning strategy fails to adequately train normal behavior models for fault detection when representative training data are lacking. The presented privacy-preserving federated learning strategy significantly improves the accuracy of normal behavior models for wind turbines lacking representative training data, as they can benefit from the training on the data of other turbines.

However, when the distributions of the monitored variable differ strongly across the fleet, a single global model shared by all turbines can deteriorate the performance of the normal behavior models, compared to conventional machine learning, even if representative training data are lacking. We have presented a customized federated learning strategy to address this challenge of heterogeneously distributed target variables. By customizing the global model to each client WT by local finetuning of neural network layers, we first successfully revert the performance losses of the global model, so that no turbine suffers from a performance loss by participating in the federated learning process. Customized federated learning yields the best model performance across all compared learning strategies. Our case studies suggest that fleet-wide learning and sharing of condition information can be achieved even where the monitored target variable is distributed heterogeneously across the fleet. Client WTs with scarce training sets were able to extract and customize knowledge from other fleet members. The federated learning process increased the average model training time by factors of 7 and 14 in the presented case studies, which can be attributed to more comprehensive communication and overhead operations and slower model convergence in the federated learning process.

Our proposed federated learning method proposes a solution to a major problem in energy and power system fleets: The lack of data sharing which "is hindering technical progress (..) in the renewable-energy industry" (Kusiak, 2016). Future research directions may involve investigating further applications of federated learning in renewable energy domains, various customization strategies, and different characteristics and effects of heterogeneously distributed target variables. It should also investigate how model training times scale with fleet size for large fleets and possibly more complex models such as multi-target normal behavior models.

**Acknowledgements**
The work of A. M. and S. J. was supported by the Swiss National Science Foundation and the Swiss Innovation agency Innosuisse.

# Appendix
## A1. Model selection
All experiments for all learning strategies and all turbines make use of the same underlying neural network hyperparameters to enable meaningful comparisons among the learning strategies and trained models in each case study. One randomly selected WT out of the ten WTs was set aside to serve as 'public' WT, with its public dataset being used only to define the network architecture with optimal hyperparameters to model the power generation and the gear bearing temperature's behaviors in normal operation. The last 30% of the SCADA data of this WT are set aside as test set, as with all other WTs. The remaining 70% are used as training data in the model search. We implemented a random search algorithm for the model search using the KerasTuner framework (O'Malley et al., 2019). In each of its trials, the algorithm randomly chooses one possible model configuration from the search space, then trains that model using the training set and finally evaluates it on the test set. The constructed model candidate is trained using the training set for up to 150 epochs or until the loss has not improved during 15 epochs. After finishing a number of 100 different trials, the hyperparameters of the trial with best performance, defined here as the lowest root mean squared error, are chosen for all further experiments. In terms of possible configurations for each trial, we have restricted the hyperparameter search space as follows:

1. Each fully-connected neural network candidate always starts with the input layer.
2. It ends with an output layer (1 unit, ReLU activation for strictly positive power, linear activation for the gear bearing temperature).
3. In between, the model can contain up to 3 hidden fully-connected layers, with each layer consisting of either 4,8, 12, or 16 units followed by an exponential linear unit (elu) activation.
4. The algorithm samples a new learning rate (between 0.075 and 0.001 in case study 1, and between 0.001 and 0.000005 in case study 2) in each trial for the stochastic gradient descent optimizer (Nesterov Momentum 0.90, batch size 32), which minimizes the mean squared error over the training set.

The best model architectures that achieve the lowest test set loss after 100 trials are described in Tables 3 and 4.

## A2. Detailed case study results

**Table A1.** Performances of the training strategies on the test set in terms of RMSE between the NBM predicted power and actual power in MW in the first case study. "Scarce" and "Repres." denote the WTs whose training sets consist of the four weeks with lowest average wind speeds and representative wind speed observations, respectively. "Conv. ML": Conventional machine learning; "Global FL": Federated learning with global model; "Cust. FL": Customized federated learning. "Training Time" is time required for the model training to finish in seconds.

| | | RMSE | | | Training Time [s] | | |
|---|---|---|---|---|---|---|---|
| WT index | Train. Dataset | Conv. ML (A) | Global FL (B) | Cust. FL (C) | Conv. ML (A) | Global FL (B) | Cust. FL (C) |
| 1 | Scarce | 0.279 | **0.110** | 0.115 | **22** | 541 | 547 |
| 2 | Scarce | 0.280 | **0.121** | 0.123 | **17** | 541 | 548 |
| 3 | Scarce | 0.200 | 0.113 | **0.087** | **19** | 541 | 549 |
| 4 | Scarce | 0.174 | **0.112** | 0.113 | **114** | 541 | 546 |
| 5 | Scarce | 0.224 | 0.168 | **0.148** | **21** | 541 | 545 |
| 6 | Repres. | **0.109** | 0.126 | 0.109 | **156** | 541 | 557 |
| 7 | Repres. | **0.106** | 0.120 | 0.106 | **117** | 541 | 561 |
| 8 | Repres. | **0.099** | 0.107 | 0.099 | **140** | 541 | 557 |
| 9 | Repres. | 0.101 | **0.100** | 0.102 | **109** | 541 | 553 |

**Table A2.** Training strategies performance on test set in terms of RMSE between the NBM predicted and actual gear bearing temperatures in °C in the second case study. "Scarce" and "Repres." denote the WTs whose training sets consist of four randomly chosen consecutive weeks and representative gear bearing temperature observations, respectively. "Conv. ML": Conventional non-collaborative machine learning; "Global FL": Federated learning with global model; "Cust. FL": Customized federated learning. "Training Time" is time required for the model training to finish in seconds.

| WT index | Train. Dataset | RMSE | | | Training Time [s] | | |
|---|---|---|---|---|---|---|---|
| | | Conv. ML (A) | Global FL (B) | Cust. FL (C) | Conv. ML (A) | Global FL (B) | Cust. FL (C) |
| 1 | Scarce | 3.85 | 3.95 | **3.79** | 19 | 627 | 641 |
| 2 | Scarce | 5.78 | 6.39 | **5.61** | 15 | 627 | 632 |
| 3 | Scarce | 3.85 | 3.59 | **3.54** | 25 | 627 | 632 |
| 4 | Scarce | 3.77 | 3.85 | **3.65** | 17 | 627 | 643 |
| 5 | Scarce | 4.19 | 3.92 | **3.91** | 14 | 627 | 643 |
| 6 | Repres. | 3.90 | 4.28 | **3.86** | 46 | 627 | 643 |
| 7 | Repres. | 3.77 | 3.87 | **3.74** | 133 | 627 | 647 |
| 8 | Repres. | **4.43** | 4.82 | 4.45 | 52 | 627 | 641 |
| 9 | Repres. | **3.62** | 3.68 | **3.62** | 65 | 627 | 648 |

## A3. Customized federated learning

We employed a customization approach by finetuning the global federated learning model, as outlined in section 4. This finetuning process, resembling a transfer learning approach, involves maintaining the weights of chosen layers and only training the weights of the remaining layers for several epochs with a smaller learning rate to adjust the pretrained weights to the local dataset of the client. The model consists of three layers with trainable weights (Table 3 and 4). Thus, we have evaluated the options of
1) only finetuning the last layer (1 finetuned layer),
2) finetuning the last two layers (2 finetuned layers), and
3) finetuning all trainable layers (3 finetuned layers),

while maintaining the weights of the other layers in accordance with their states in the global federated learning model. For each of the three options, we trained the model using a smaller learning rate (half of the learning rate used in the conventional and standard federated learning process) until the validation loss, defined as the root mean squared error on the validation set, did not improve for 5 epochs. Tables A3 and A4 show the results on the validation set for each client WT in the case studies. For each client turbine, we choose the best performing model from the three options, that is, the model with the lowest validation loss, as the customized federated learning model used for the evaluation in Tables A1 and A2.

**Table A3.** The root mean squared errors calculated over the respective client WT's validation set of three evaluated customization experiments in case study 1. "WT": wind turbine, "FL": federated learning.

| WT index | Train Dataset | Customized FL, 1 layer | Customized FL, 2 layers | Customized FL, 3 layers |
|---|---|---|---|---|
| 1 | Scarce | .0954 | .0954 | **.0947** |
| 2 | Scarce | .1043 | .1066 | **.1026** |
| 3 | Scarce | .0832 | .0825 | **.0809** |
| 4 | Scarce | .1032 | .1038 | **.1032** |
| 5 | Scarce | .1466 | .1473 | **.1424** |
| 6 | Repres. | .0958 | **.0956** | .0959 |
| 7 | Repres. | .0864 | **.0859** | .0861 |
| 8 | Repres. | **.0806** | .0807 | .0807 |
| 9 | Repres. | .0866 | **.0864** | .0866 |

**Table A4.** The root mean squared errors calculated over the respective client WT's validation set of three evaluated customization experiments in case study 2. "WT": wind turbine, "FL": federated learning.

| WT index | Train Dataset | Customized FL, 1 layer | Customized FL, 2 layers | Customized FL, 3 layers |
|---|---|---|---|---|
| 1 | Scarce | **3.656** | 4.126 | 3.921 |
| 2 | Scarce | **4.942** | 5.066 | 4.960 |
| 3 | Scarce | **3.678** | 3.767 | 3.779 |
| 4 | Scarce | **3.702** | 3.952 | 3.863 |
| 5 | Scarce | **4.676** | 4.712 | 4.704 |
| 6 | Repres. | **3.710** | 3.713 | 3.720 |
| 7 | Repres. | **3.733** | 3.735 | 3.745 |
| 8 | Repres. | **5.670** | 5.720 | 5.697 |
| 9 | Repres. | **3.809** | 3.829 | 3.860 |

### A4. Second wind farm dataset

We additionally investigated our two presented case studies (section 3) using data from the publicly available Penmanshiel wind farm dataset (Plumley, 2022). The onshore wind farm consists of 14 identical WTs of the same configuration (Table A5). The dataset comprises 10-minute averages of SCADA data recorded across a time span of 5 years. Each WT's local dataset contains around 150'000 valid datapoints per variable, which includes wind speeds measured at the nacelle, power generation, and gear bearing temperatures. We assume that no data sharing between WTs is allowed.

**Table A5.** Technical specifications of the wind turbines from the Penmanshiel wind farm employed in the case studies.

| Parameter | Specification |
|---|---|
| Rotor diameter | 82 m |
| Rated active power | 2050 kW |
| Cut-in wind velocity | 3.5 m/s |
| Cut-out wind velocity | 25 m/s |
| Tower | Steel |
| Control type | Electrical pitch system |
| Gearbox | Combined planetary/spur |

### A4.1 Case Studies

We apply the identical case study designs as outlined in sections 3.1 and 3.2. For the first case study, the federated learning of active power models, the normalized 10-minute average wind speed serves as regressor of the power generation. Seven WTs, that is 50% of the turbines in the wind farm, are affected by a lack of representative training data in our scenario. For each WT, the last 30% of its SCADA data are set aside as test set. The remaining 70% are split into a training and validation set. For the seven WTs affected by data scarcity, only the four weeks with the lowest average wind speeds comprise the training set, with the remainder belonging to the validation set. For the other half, the remaining 70% of SCADA data are split into a training set (the first 70% of data) and a validation set (the last 30% of data). Figure A1 illustrates the training, validation, and test set for one of the seven data-scarce WTs and one of the seven WTs with representative training sets.

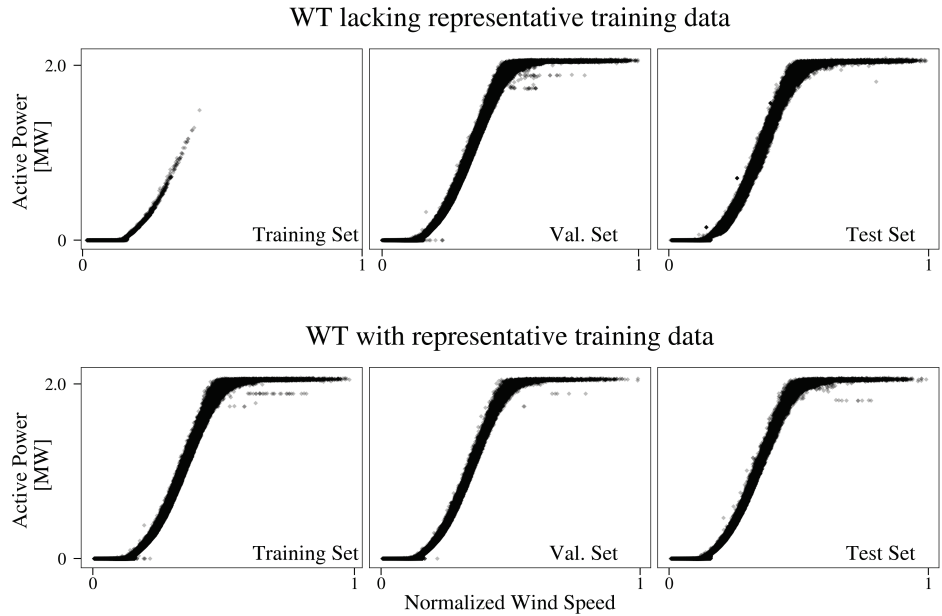

**Figure A1.** Datasets of two different client turbines from the Penmanshiel wind farm. First row: Only data from the four weeks with the lowest average wind speed were kept for the training set of this client turbine. The training set does not contain sufficient data to represent the true power curve behavior in high wind speed situations (upper left panel). Second row: Wind speed and power data from a client WT whose training data contain representatively distributed wind speed observations.

For the second case study, the federated learning of bearing temperature models, the normalized 10-minute rotor speeds and power generation are regressor inputs to the model predicting the (front) bearing temperature. The WTs' datasets were split according to the same scheme as in the first case study, with the difference that the training sets of the seven randomly chosen WTs affected by the data scarcity scenario now consist of only randomly selected four consecutive weeks of data. Figure A2 illustrates the training, validation, and test set for one of the seven data-scarce WTs and one of the seven WTs with representative training sets.

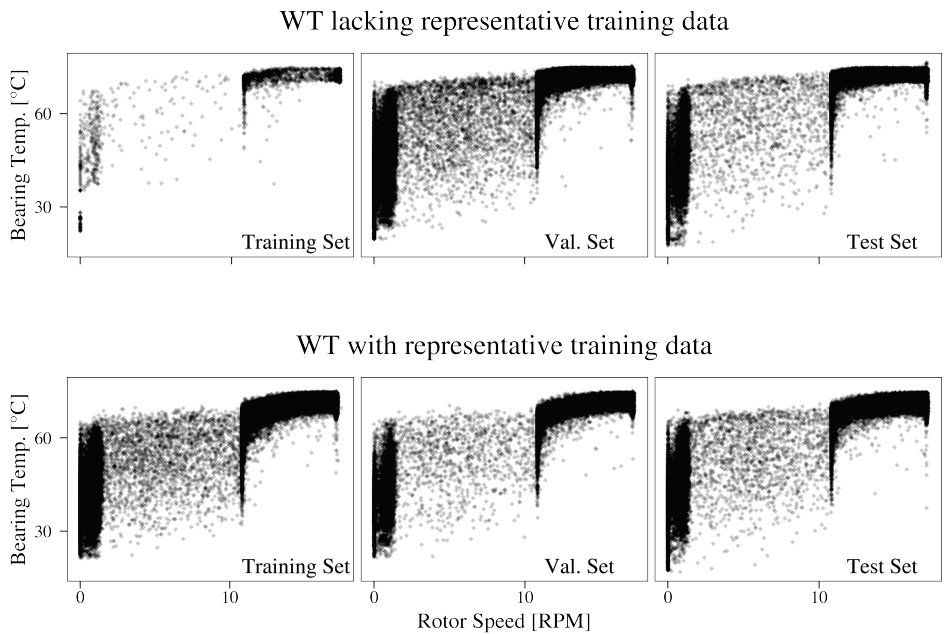

**Figure A2.** Datasets of two client WTs from the Penmanshiel wind farm. First row: Only data from four randomly chosen consecutive weeks were kept for the training set of this client turbine. Second row: Gear bearing temperature and rotor speed data from a client WT whose training data contain representatively temperature observations.

Figure A3 shows the distributions of the monitored variables in each case study (power generation, bearing temperature) for all 14 WTs in the wind farm. While the distributions of the active power exhibit almost identical distributions across the wind farm, the temperature distributions show significant differences. These characteristics are in accordance with the discussed setting in section 3.3.

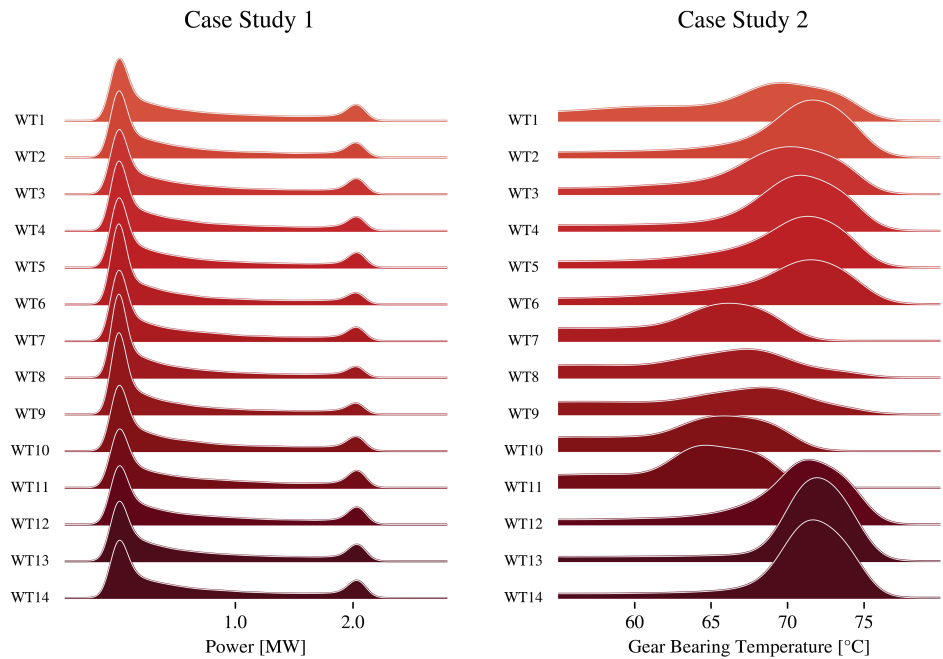

**Figure A3.** Kernel density estimates of the distributions of the monitored variables on the test set of all nine client WTs.

### A4.2 Results
We evaluate the presented strategies A-C (conventional machine learning, global federated learning, customized federated learning) from section 4.1 for both case studies.

### A4.2.1 Federated Learning of Active Power models
In case study 1, a NBM of the power generation is trained. We use the same model architecture and configuration summarized in Table 4. The results, shown in Figure A4 and Table A6, validate our previous findings of case study 1 discussed in section 4.2. For WTs lacking representative training data, a conventional machine learning strategy results in a poor fit (mean RMSE: 0.188), as the local training sets are lacking representative data for high wind speed ranges. These WTs benefit from a significant error reduction by participating in the global federated learning process (mean: 0.039). The global model however results in a performance loss for WTs with representative training sets (mean: 0.038) compared to strategy A (mean: 0.035). Customized federated learning reverts these performance losses back to the original level (mean: 0.034) by enabling the WTs to adjust the global model to their local datasets, thus resulting again in overall the best performing strategy. In terms of computational time, the average training time of the federated learning strategy increased by a factor of 18 compared to the conventional machine learning strategy. The additional training time for the customized federated learning strategy, i.e., the local finetuning, remains negligible (on average +29.4 seconds).

Table A6. Performances of the training strategies on the test set in terms of RMSE between the NBM predicted power and actual power in MW in the first case study with data from the Penmanshiel wind farm. "Scarce" and "Repres." denote the WTs whose training sets consist of the four weeks with lowest average wind speeds and representative wind speed observations, respectively. "Conv. ML": Conventional machine learning; "Global FL": Federated learning with global model; "Cust. FL": Customized federated learning. "Training Time" is time required for the model training to finish in seconds.

| WT index | Train. Dataset | RMSE | | | Training Time [s] | | |
|---|---|---|---|---|---|---|---|
| | | Conv. ML (A) | Global FL (B) | Cust. FL (C) | Conv. ML (A) | Global FL (B) | Cust. FL (C) |
| 1 | Scarce | 0.182 | 0.032 | **0.031** | 127 | 1680 | 1688 |
| 2 | Scarce | 0.196 | 0.055 | **0.043** | 37 | 1680 | 1687 |
| 3 | Scarce | 0.191 | 0.033 | **0.033** | 40 | 1680 | 1688 |
| 4 | Scarce | 0.166 | 0.035 | **0.032** | 48 | 1680 | 1687 |
| 5 | Scarce | 0.240 | **0.031** | 0.031 | 185 | 1680 | 1687 |
| 6 | Scarce | 0.174 | **0.030** | 0.030 | 51 | 1680 | 1689 |
| 7 | Scarce | 0.172 | **0.057** | 0.058 | 79 | 1680 | 1688 |
| 8 | Repres. | 0.032 | 0.031 | **0.031** | 71 | 1680 | 1715 |
| 9 | Repres. | **0.031** | 0.039 | 0.031 | 130 | 1680 | 1737 |
| 10 | Repres. | 0.034 | **0.033** | 0.033 | 78 | 1680 | 1707 |
| 11 | Repres. | **0.040** | 0.048 | 0.040 | 84 | 1680 | 1714 |
| 12 | Repres. | 0.034 | **0.033** | 0.033 | 172 | 1680 | 1742 |
| 13 | Repres. | 0.033 | **0.032** | 0.032 | 81 | 1680 | 1763 |
| 14 | Repres. | 0.041 | 0.052 | **0.040** | 95 | 1680 | 1735 |

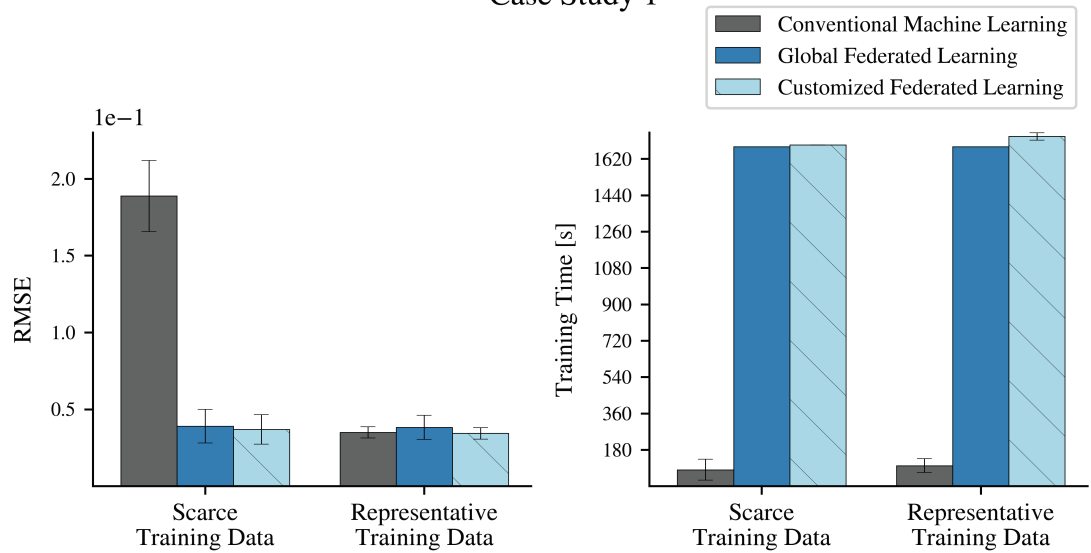

**Figure A4.** Left: Performances of the training strategies on the test set in terms of mean RMSE between the NBM predicted power and actual monitored variable in case study 1. Right: Mean training time in seconds for all three learning strategies. The error bars display the standard deviation.

### A4.2.2 Federated Learning of Bearing Temperature models

In case study 2, a NBM of the bearing temperature is trained. We employ the identical model architecture and configuration as summarized in Table 4. The results, shown in Figure A5 and Table A7, also validate our previous findings of case study 2 discussed in section 4.3. A global federated learning strategy results in a significant performance deterioration, even for the seven WTs lacking representative training data (increase in mean RMSE from 6.10 to 7.21). These results further suggest that the strongly deviating bearing temperature distributions across the fleet can negatively impact the generalizability of the global model, resulting in an inadequate fit to most participants. Consistent with our observations from section 4.3, a customized federated learning strategy can not only revert the performance losses for WTs with representative training data (mean RMSE by strategies: A: 5.82, B: 7.04, C: 5.82), it also enables data-scarce WTs to retain and transfer knowledge from the global model, such that this strategy results in the lowest error for these WTs in this scenario (mean: 5.91). The average training time of the global federated learning strategy increased by a factor of 7 compared to the conventional machine learning strategy, while the efficient local finetuning step only required an average additional training time of 30.9 seconds.

**Table A7.** Training strategies performance on test set in terms of RMSE between the NBM predicted and actual gear bearing temperatures in °C in the second case study using data from the Penmanshield wind farm dataset. "Scarce" and "Repres." denote the WTs whose training sets consist of four randomly chosen consecutive weeks and representative gear bearing temperature observations, respectively. "Conv. ML": Conventional non-collaborative machine learning; "Global FL": Federated learning with global model; "Cust. FL": Customized federated learning. "Training Time" is time required for the model training to finish in seconds.

| WT index | Train. Dataset | RMSE Conv. ML (A) | RMSE Global FL (B) | RMSE Cust. FL (C) | Training Time [s] Conv. ML (A) | Training Time [s] Global FL (B) | Training Time [s] Cust. FL (C) |
|---|---|---|---|---|---|---|---|
| 1 | Scarce | 5.857 | 5.413 | **5.407** | **10** | 588 | 591 |
| 2 | Scarce | 7.397 | 7.323 | **6.978** | **26** | 588 | 591 |
| 3 | Scarce | 6.146 | 7.810 | **6.131** | **35** | 588 | 598 |
| 4 | Scarce | **5.310** | 7.254 | 5.313 | **45** | 588 | 594 |
| 5 | Scarce | 5.918 | 6.968 | **5.700** | **23** | 588 | 599 |
| 6 | Scarce | 6.056 | 8.035 | **5.941** | **48** | 588 | 597 |
| 7 | Scarce | 6.021 | 7.667 | **5.888** | **18** | 588 | 593 |
| 8 | Repres. | **5.980** | 6.970 | 6.061 | **121** | 588 | 624 |
| 9 | Repres. | 5.962 | 6.503 | **5.910** | **100** | 588 | 664 |
| 10 | Repres. | **5.981** | 7.049 | 6.001 | **207** | 588 | 651 |
| 11 | Repres. | **6.094** | 6.780 | 6.108 | **115** | 588 | 611 |
| 12 | Repres. | **5.649** | 7.400 | 5.656 | **127** | 588 | 680 |
| 13 | Repres. | 6.128 | 7.826 | **6.013** | **142** | 588 | 637 |
| 14 | Repres. | **4.964** | 6.778 | 4.995 | **123** | 588 | 635 |

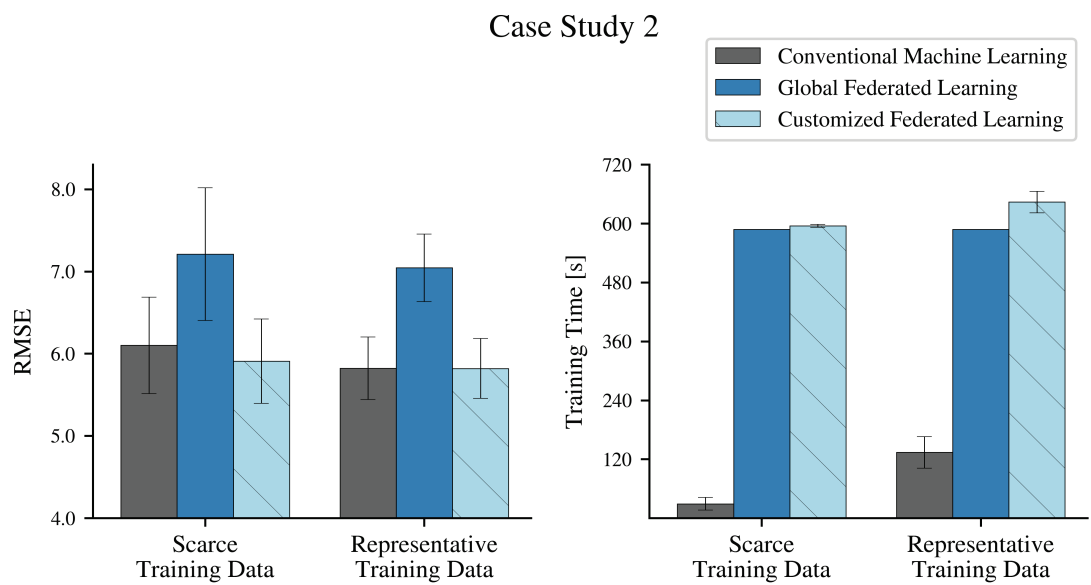

**Figure A5.** Left: Performances of the training strategies on the test set in terms of mean RMSE between the NBM predicted temperature and the actual monitored variable in case study 2. Right: Mean training time in seconds for all three learning strategies. The error bars display the standard deviation.

# References

Abadi, M., Agarwal, A., Barham, P., Brevdo, E., Chen, Z., Citro, C., Corrado, G. S., Davis, A., Dean, J., Devin, M., Ghemawat, S., Goodfellow, I., Harp, A., Irving, G., Isard, M., Jia, Y., Jozefowicz, R., Kaiser, L., Kudlur, M., … Zheng, X. (2016). *TensorFlow: Large-Scale Machine Learning on Heterogeneous Distributed Systems*. https://doi.org/10.48550/ARXIV.1603.04467

Acar, D. A. E., Zhao, Y., Matas, R., Mattina, M., Whatmough, P., & Saligrama, V. (2021). Federated Learning Based on Dynamic Regularization. *International Conference on Learning Representations*. https://openreview.net/forum?id=B7v4QMR6Z9w

Aledhari, M., Razzak, R., Parizi, R. M., & Saeed, F. (2020). Federated Learning: A Survey on Enabling Technologies, Protocols, and Applications. *IEEE Access*, *8*, 140699–140725. https://doi.org/10.1109/ACCESS.2020.3013541

Arivazhagan, M. G., Aggarwal, V., Singh, A. K., & Choudhary, S. (2019). *Federated Learning with Personalization Layers*. https://doi.org/10.48550/ARXIV.1912.00818

Asad, M., Moustafa, A., & Ito, T. (2020). FedOpt: Towards Communication Efficiency and Privacy Preservation in Federated Learning. *Applied Sciences*, *10*(8), 2864. https://doi.org/10.3390/app10082864

Badihi, H., Zhang, Y., Jiang, B., Pillay, P., & Rakheja, S. (2022). A Comprehensive Review on Signal-Based and Model-Based Condition Monitoring of Wind Turbines: Fault Diagnosis and Lifetime Prognosis. *Proceedings of the IEEE*, *110*(6), 754–806. https://doi.org/10.1109/JPROC.2022.3171691

Barthelmie, R. J., & Pryor, S. C. (2021). Climate Change Mitigation Potential of Wind Energy. *Climate*, *9*(9), 136. https://doi.org/10.3390/cli9090136

Bilendo, F., Badihi, H., Lu, N., Cambron, P., & Jiang, B. (2022). A Normal Behavior Model Based on Power Curve and Stacked Regressions for Condition Monitoring of Wind Turbines. *IEEE Transactions on Instrumentation and Measurement*, *71*, 1–13. https://doi.org/10.1109/TIM.2022.3196116

Bilendo, F., Meyer, A., Badihi, H., Lu, N., Cambron, P., & Jiang, B. (2022). Applications and Modeling Techniques of Wind Turbine Power Curve for Wind Farms—A Review. *Energies*, *16*(1), 180. https://doi.org/10.3390/en16010180

Black, I. M., Richmond, M., & Kolios, A. (2021). Condition monitoring systems: A systematic literature review on machine-learning methods improving offshore-wind turbine operational management. *International Journal of Sustainable Energy*, *40*(10), 923–946. https://doi.org/10.1080/14786451.2021.1890736

Carroll, J., McDonald, A., & McMillan, D. (2016). Failure rate, repair time and unscheduled O&M cost analysis of offshore wind turbines: Reliability and maintenance of offshore wind turbines. *Wind Energy*, *19*(6), 1107–1119. https://doi.org/10.1002/we.1887

Cheng, X., Tian, W., Shi, F., Zhao, M., Chen, S., & Wang, H. (2022). A Blockchain-Empowered Cluster-Based Federated Learning Model for Blade Icing Estimation on IoT-Enabled Wind Turbine. *IEEE Transactions on Industrial Informatics*, *18*(12), 9184–9195. https://doi.org/10.1109/TII.2022.3159684

Chollet, F. & others. (2015). *Keras*. https://keras.io

Clifton, A., Barber, S., Bray, A., Enevoldsen, P., Fields, J., Sempreviva, A. M., Williams, L., Quick, J., Purdue, M., Totaro, P., & Ding, Y. (2022). *Grand Challenges in the Digitalisation of Wind Energy* [Preprint]. Operation, condition monitoring, and maintenance. https://doi.org/10.5194/wes-2022-29

Collins, L., Hassani, H., Mokhtari, A., & Shakkottai, S. (2022). *FedAvg with Fine Tuning: Local Updates Lead to Representation Learning*. https://doi.org/10.48550/ARXIV.2205.13692

Dao, P. B. (2022). On Wilcoxon rank sum test for condition monitoring and fault detection of wind turbines. *Applied Energy*, *318*, 119209. https://doi.org/10.1016/j.apenergy.2022.119209

Edenhofer, O., Pichs-Madruga, R., Sokona, Y., Seyboth, K., Kadner, S., Zwickel, T., Eickemeier, P., Hansen, G., Schlömer, S., Von Stechow, C., & Matschoss, P. (Eds.). (2011). *Renewable Energy Sources and Climate Change Mitigation: Special Report of the Intergovernmental Panel on Climate Change* (1st ed.). Cambridge University Press. https://doi.org/10.1017/CBO9781139151153

Faulstich, S., Hahn, B., & Tavner, P. J. (2011). Wind turbine downtime and its importance for offshore deployment. *Wind Energy*, *14*(3), 327–337. https://doi.org/10.1002/we.421

García Márquez, F. P., Tobias, A. M., Pinar Pérez, J. M., & Papaelias, M. (2012). Condition monitoring of wind turbines: Techniques and methods. *Renewable Energy*, *46*, 169–178. https://doi.org/10.1016/j.renene.2012.03.003

Hard, A., Rao, K., Mathews, R., Ramaswamy, S., Beaufays, F., Augenstein, S., Eichner, H., Kiddon, C., & Ramage, D. (2018). *Federated Learning for Mobile Keyboard Prediction*. https://doi.org/10.48550/ARXIV.1811.03604

Huang, W., Ye, M., Shi, Z., Li, H., & Du, B. (2023). Rethinking Federated Learning With Domain Shift: A Prototype View. *Proceedings of the IEEE/CVF Conference on Computer Vision and Pattern Recognition (CVPR)*, 16312–16322.

IEA. (2021). *Renewables 2021*. IEA. https://www.iea.org/reports/renewables-2021

IEA. (2022). *World Energy Investment 2022*. IEA. https://www.iea.org/reports/world-energy-investment-2022

Jonas, S., Anagnostos, D., Brodbeck, B., & Meyer, A. (2023). Vibration Fault Detection in Wind Turbines Based on Normal Behaviour Models without Feature Engineering. *Energies*, *16*(4), 1760. https://doi.org/10.3390/en16041760

Kairouz, P., McMahan, H. B., Avent, B., Bellet, A., Bennis, M., Nitin Bhagoji, A., Bonawitz, K., Charles, Z., Cormode, G., Cummings, R., D'Oliveira, R. G. L., Eichner, H., El Rouayheb, S., Evans, D., Gardner, J., Garrett, Z., Gascón, A., Ghazi, B., Gibbons, P. B., … Zhao, S. (2021). Advances and Open Problems in Federated Learning. *Foundations and Trends® in Machine Learning*, *14*(1–2), 1–210. https://doi.org/10.1561/2200000083

Kouw, W. M., & Loog, M. (2018). *An introduction to domain adaptation and transfer learning*. https://doi.org/10.48550/ARXIV.1812.11806

Kulkarni, V., Kulkarni, M., & Pant, A. (2020). Survey of Personalization Techniques for Federated Learning. *2020 Fourth World Conference on Smart Trends in Systems, Security and Sustainability (WorldS4)*, 794–797. https://doi.org/10.1109/WorldS450073.2020.9210355

Kusiak, A. (2016). Renewables: Share data on wind energy. *Nature*, *529*(7584), 19–21. https://doi.org/10.1038/529019a

Kusiak, A., Zheng, H., & Song, Z. (2009). On-line monitoring of power curves. *Renewable Energy*, *34*(6), 1487–1493. https://doi.org/10.1016/j.renene.2008.10.022

Leahy, K., Gallagher, C., O'Donovan, P., & O'Sullivan, D. T. J. (2019). Issues with Data Quality for Wind Turbine Condition Monitoring and Reliability Analyses. *Energies*, *12*(2), 201. https://doi.org/10.3390/en12020201

Li, L., Fan, Y., Tse, M., & Lin, K.-Y. (2020). A review of applications in federated learning. *Computers & Industrial Engineering*, *149*, 106854. https://doi.org/10.1016/j.cie.2020.106854

Li, Q., Diao, Y., Chen, Q., & He, B. (2022). Federated Learning on Non-IID Data Silos: An Experimental Study. *2022 IEEE 38th International Conference on Data Engineering (ICDE)*, 965–978. https://doi.org/10.1109/ICDE53745.2022.00077

Li, T., Sahu, A. K., Talwalkar, A., & Smith, V. (2020). Federated Learning: Challenges, Methods, and Future Directions. *IEEE Signal Processing Magazine*, *37*(3), 50–60. https://doi.org/10.1109/MSP.2020.2975749

Lim, W. Y. B., Luong, N. C., Hoang, D. T., Jiao, Y., Liang, Y.-C., Yang, Q., Niyato, D., & Miao, C. (2020). Federated Learning in Mobile Edge Networks: A Comprehensive Survey. *IEEE Communications Surveys & Tutorials*, *22*(3), 2031–2063. https://doi.org/10.1109/COMST.2020.2986024

Lin, J., Ma, J., & Zhu, J. (2022). A Privacy-Preserving Federated Learning Method for Probabilistic Community-Level Behind-the-Meter Solar Generation Disaggregation. *IEEE Transactions on Smart Grid*, *13*(1), 268–279. https://doi.org/10.1109/TSG.2021.3115904

Liu, Q., Yang, B., Wang, Z., Zhu, D., Wang, X., Ma, K., & Guan, X. (2022). Asynchronous Decentralized Federated Learning for Collaborative Fault Diagnosis of PV Stations. *IEEE Transactions on Network Science and Engineering*, *9*(3), 1680–1696. https://doi.org/10.1109/TNSE.2022.3150182

Liu, Y., Yu, J. J. Q., Kang, J., Niyato, D., & Zhang, S. (2020). Privacy-Preserving Traffic Flow Prediction: A Federated Learning Approach. *IEEE Internet of Things Journal*, *7*(8), 7751–7763. https://doi.org/10.1109/JIOT.2020.2991401

Lu, Y., Huang, X., Zhang, K., Maharjan, S., & Zhang, Y. (2020). Blockchain Empowered Asynchronous Federated Learning for Secure Data Sharing in Internet of Vehicles. *IEEE Transactions on Vehicular Technology*, *69*(4), 4298–4311. https://doi.org/10.1109/TVT.2020.2973651

Lydia, M., Kumar, S. S., Selvakumar, A. I., & Prem Kumar, G. E. (2014). A comprehensive review on wind turbine power curve modeling techniques. *Renewable and Sustainable Energy Reviews*, *30*, 452–460. https://doi.org/10.1016/j.rser.2013.10.030

Marvuglia, A., & Messineo, A. (2012). Monitoring of wind farms' power curves using machine learning techniques. *Applied Energy*, *98*, 574–583. https://doi.org/10.1016/j.apenergy.2012.04.037

McMahan, B., Moore, E., Ramage, D., Hampson, S., & Arcas, B. A. y. (2017). Communication-Efficient Learning of Deep Networks from Decentralized Data. In A. Singh & J. Zhu (Eds.), *Proceedings of the 20th International Conference on Artificial Intelligence and Statistics* (Vol. 54, pp. 1273–1282). PMLR. https://proceedings.mlr.press/v54/mcmahan17a.html

Meyer, A. (2021). Multi-target normal behaviour models for wind farm condition monitoring. *Applied Energy*, *300*, 117342. https://doi.org/10.1016/j.apenergy.2021.117342

Meyer, A., & Brodbeck, B. (2020). Data-driven performance fault detection in commercial wind turbines. *PHM Society European Conference*, *5*, 7–7.

Mothukuri, V., Parizi, R. M., Pouriyeh, S., Huang, Y., Dehghantanha, A., & Srivastava, G. (2021). A survey on security and privacy of federated learning. *Future Generation Computer Systems*, *115*, 619–640. https://doi.org/10.1016/j.future.2020.10.007

Nair, V., & Hinton, G. E. (2010). Rectified linear units improve restricted boltzmann machines. *Proceedings of the 27th International Conference on Machine Learning (ICML-10)*, 807–814.

Nunes, A. R., Morais, H., & Sardinha, A. (2021). Use of Learning Mechanisms to Improve the Condition Monitoring of Wind Turbine Generators: A Review. *Energies*, *14*(21), 7129. https://doi.org/10.3390/en14217129

OECD, The World Bank, & United Nations Environment Programme. (2018). *Financing Climate Futures: Rethinking Infrastructure*. OECD. https://doi.org/10.1787/9789264308114-en

Ohlendorf, N., & Schill, W.-P. (2020). Frequency and duration of low-wind-power events in Germany. *Environmental Research Letters*, *15*(8), 084045. https://doi.org/10.1088/1748-9326/ab91e9

O'Malley, T., Bursztein, E., Long, J., Chollet, F., Jin, H., Invernizzi, L., & others. (2019). *KerasTuner*. https://github.com/keras-team/keras-tuner

Pan, S. J., & Yang, Q. (2010). A Survey on Transfer Learning. *IEEE Transactions on Knowledge and Data Engineering*, *22*(10), 1345–1359. https://doi.org/10.1109/TKDE.2009.191

Pandit, R., Astolfi, D., Hong, J., Infield, D., & Santos, M. (2023). SCADA data for wind turbine data-driven condition/performance monitoring: A review on state-of-art, challenges and future trends. *Wind Engineering*, *47*(2), 422–441. https://doi.org/10.1177/0309524X221124031

Pichai, S. (2019). Google's Sundar Pichai: Privacy Should Not Be a Luxury Good. *The New York Times*. https://www.nytimes.com/2019/05/07/opinion/google-sundar-pichai-privacy.html

Pinar Pérez, J. M., García Márquez, F. P., Tobias, A., & Papaelias, M. (2013). Wind turbine reliability analysis. *Renewable and Sustainable Energy Reviews*, *23*, 463–472. https://doi.org/10.1016/j.rser.2013.03.018

Plumley, C. (2022). *Penmanshiel Wind Farm Data* (0.0.2) [Data set]. Zenodo. https://doi.org/10.5281/ZENODO.5946808

Quinonero-Candela, J., Sugiyama, M., Schwaighofer, A., & Lawrence, N. D. (2008). *Dataset shift in machine learning*. Mit Press.

Schlechtingen, M., Santos, I. F., & Achiche, S. (2013a). Using Data-Mining Approaches for Wind Turbine Power Curve Monitoring: A Comparative Study. *IEEE Transactions on Sustainable Energy*, *4*(3), 671–679. https://doi.org/10.1109/TSTE.2013.2241797

Schlechtingen, M., Santos, I. F., & Achiche, S. (2013b). Wind turbine condition monitoring based on SCADA data using normal behavior models. Part 1: System description. *Applied Soft Computing*, *13*(1), 259–270. https://doi.org/10.1016/j.asoc.2012.08.033

Shamsian, A., Navon, A., Fetaya, E., & Chechik, G. (2021). Personalized federated learning using hypernetworks. *International Conference on Machine Learning*, 9489–9502.

Shokrzadeh, S., Jafari Jozani, M., & Bibeau, E. (2014). Wind Turbine Power Curve Modeling Using Advanced Parametric and Nonparametric Methods. *IEEE Transactions on Sustainable Energy*, *5*(4), 1262–1269. https://doi.org/10.1109/TSTE.2014.2345059

Siemens Gamesa. (2022). *The power of big data*. https://www.siemensgamesa.com/explore/innovations/digitalization

Smith, V., Chiang, C.-K., Sanjabi, M., & Talwalkar, A. S. (2017). Federated Multi-Task Learning. In I. Guyon, U. V. Luxburg, S. Bengio, H. Wallach, R. Fergus, S. Vishwanathan, & R. Garnett (Eds.), *Advances in Neural Information Processing Systems* (Vol. 30). Curran Associates, Inc. https://proceedings.neurips.cc/paper_files/paper/2017/file/6211080fa89981f66b1a0c9d55c61d0f-Paper.pdf

Stetco, A., Dinmohammadi, F., Zhao, X., Robu, V., Flynn, D., Barnes, M., Keane, J., & Nenadic, G. (2019). Machine learning methods for wind turbine condition monitoring: A review. *Renewable Energy*, *133*, 620–635. https://doi.org/10.1016/j.renene.2018.10.047

Sun, S., Wang, T., Yang, H., & Chu, F. (2022). Condition monitoring of wind turbine blades based on self-supervised health representation learning: A conducive technique to effective and reliable utilization of wind energy. *Applied Energy*, *313*, 118882. https://doi.org/10.1016/j.apenergy.2022.118882

Tan, A. Z., Yu, H., Cui, L., & Yang, Q. (2022). Towards Personalized Federated Learning. *IEEE Transactions on Neural Networks and Learning Systems*, 1–17. https://doi.org/10.1109/TNNLS.2022.3160699

Tautz-Weinert, J., & Watson, S. J. (2017). Using SCADA data for wind turbine condition monitoring – a review. *IET Renewable Power Generation*, *11*(4), 382–394. https://doi.org/10.1049/iet-rpg.2016.0248

Tchakoua, P., Wamkeue, R., Ouhrouche, M., Slaoui-Hasnaoui, F., Tameghe, T., & Ekemb, G. (2014). Wind Turbine Condition Monitoring: State-of-the-Art Review, New Trends, and Future Challenges. *Energies*, *7*(4), 2595–2630. https://doi.org/10.3390/en7042595

Thorgeirsson, A. T., Scheubner, S., Funfgeld, S., & Gauterin, F. (2021). Probabilistic Prediction of Energy Demand and Driving Range for Electric Vehicles With Federated Learning. *IEEE Open Journal of Vehicular Technology*, *2*, 151–161. https://doi.org/10.1109/OJVT.2021.3065529

Wang, A., Pei, Y., Qian, Z., Zareipour, H., Jing, B., & An, J. (2022). A two-stage anomaly decomposition scheme based on multi-variable correlation extraction for wind turbine fault detection and identification. *Applied Energy*, *321*, 119373. https://doi.org/10.1016/j.apenergy.2022.119373

Wang, Y., Hu, Q., Li, L., Foley, A. M., & Srinivasan, D. (2019). Approaches to wind power curve modeling: A review and discussion. *Renewable and Sustainable Energy Reviews*, *116*, 109422. https://doi.org/10.1016/j.rser.2019.109422

Wymore, M. L., Van Dam, J. E., Ceylan, H., & Qiao, D. (2015). A survey of health monitoring systems for wind turbines. *Renewable and Sustainable Energy Reviews*, *52*, 976–990. https://doi.org/10.1016/j.rser.2015.07.110

Yang, Q., Liu, Y., Chen, T., & Tong, Y. (2019). Federated Machine Learning: Concept and Applications. *ACM Transactions on Intelligent Systems and Technology*, *10*(2), 1–19. https://doi.org/10.1145/3298981

Yin, X., Zhu, Y., & Hu, J. (2022). A Comprehensive Survey of Privacy-preserving Federated Learning: A Taxonomy, Review, and Future Directions. *ACM Computing Surveys*, *54*(6), 1–36. https://doi.org/10.1145/3460427

Zaher, A., McArthur, S. D. J., Infield, D. G., & Patel, Y. (2009). Online wind turbine fault detection through automated SCADA data analysis. *Wind Energy*, *12*(6), 574–593. https://doi.org/10.1002/we.319

Zhang, X., Fang, F., & Wang, J. (2021). Probabilistic Solar Irradiation Forecasting Based on Variational Bayesian Inference With Secure Federated Learning. *IEEE Transactions on Industrial Informatics*, *17*(11), 7849–7859. https://doi.org/10.1109/TII.2020.3035807

Zhao, Y., Li, M., Lai, L., Suda, N., Civin, D., & Chandra, V. (2018). *Federated Learning with Non-IID Data*. https://doi.org/10.48550/ARXIV.1806.00582

Zhu, H., Xu, J., Liu, S., & Jin, Y. (2021). Federated learning on non-IID data: A survey. *Neurocomputing*, *465*, 371–390. https://doi.org/10.1016/j.neucom.2021.07.098

Zhu, Y., Zhu, C., Tan, J., Tan, Y., & Rao, L. (2022). Anomaly detection and condition monitoring of wind turbine gearbox based on LSTM-FS and transfer learning. *Renewable Energy*, *189*, 90–103. https://doi.org/10.1016/j.renene.2022.02.061

Zhuang, F., Qi, Z., Duan, K., Xi, D., Zhu, Y., Zhu, H., Xiong, H., & He, Q. (2021). A Comprehensive Survey on Transfer Learning. *Proceedings of the IEEE*, *109*(1), 43–76. https://doi.org/10.1109/JPROC.2020.3004555